\documentclass[journal]{IEEEtran}
\usepackage{cite}

\usepackage[pdftex]{graphicx}
\ifCLASSINFOpdf
   
\else
\fi

\usepackage{subfigure,balance}
\usepackage{color}
\def \etal {et al.}
\usepackage{booktabs}
\usepackage{mathtools}
\usepackage{enumitem}
\usepackage{amsmath,amsfonts}
\usepackage{multirow}

\begin{document}

%
\title{Driving Anomaly Detection Using Conditional Generative Adversarial Network}
%
%
%

\author{Yuning~Qiu,~\IEEEmembership{Student Member,~IEEE,}
        Teruhisa~Misu,~\IEEEmembership{Member,~IEEE,}
        and~Carlos~Busso,~\IEEEmembership{Senior Member,~IEEE}
\thanks{Y. Qiu and C. Busso are with the Department of Electrical Engineering, University of Texas at Dallas,
        Richardson, Texas, 75080 USA e-mail: {\tt\small \{yxq180000, busso\}@utdallas.edu}}
\thanks{Teruhisa Misu is with the Honda Research Institute,
        Mountain View, California, US
        e-mail: \tt\small \ TMisu@hra.com}}

%
%

\markboth{}%
{Shell \MakeLowercase{\textit{et al.}}: Bare Demo of IEEEtran.cls for IEEE Journals}
%



\maketitle

\begin{abstract}

Anomaly driving detection is an important problem in \emph{advanced driver assistance systems}  (ADAS). It is important to identify potential hazard scenarios as early as possible to avoid potential accidents. This study proposes an unsupervised method to quantify driving anomalies using a conditional \emph{generative adversarial network} (GAN). The approach predicts upcoming driving scenarios by conditioning the models on the previously observed signals. The system uses the difference of the output from the discriminator between the predicted and actual signals as a metric to quantify the anomaly degree of a driving segment. We take a driver-centric approach, considering physiological signals from the driver and \emph{controller area network}-Bus (CAN-Bus) signals from the vehicle. The approach is implemented with \emph{convolutional neural networks} (CNNs) to extract discriminative feature representations, and with \emph{long short-term memory} (LSTM) cells to capture temporal information. The study is implemented and evaluated with the \emph{driving anomaly dataset} (DAD), which includes 250 hours of naturalistic recordings manually annotated with driving events. The experimental results reveal that recordings annotated with events that are likely to be anomalous, such as avoiding on-road pedestrians and traffic rule violations, have higher anomaly scores than recordings without any event annotation. The results are validated with perceptual evaluations, where annotators are asked to assess the risk and familiarity of the videos detected with high anomaly scores. The results indicate that the driving segments with higher anomaly scores are more risky and less regularly seen on the road than other driving segments, validating the proposed unsupervised approach. 

\end{abstract}

\begin{IEEEkeywords}
Driving anomaly detection, conditional generative adversarial networks, convolutional neural networks, long short-term memory cell.
\end{IEEEkeywords}

%
\IEEEpeerreviewmaketitle

\section{Introduction}
%
%
%
%

\IEEEPARstart{W}{ith} the development of the smart automobile industry in recent years, more and more functions have been added to \emph{advanced driver assistance systems} (ADAS), avoiding human errors and increasing road safety. Examples include \emph{lane departure warning} (LDW), \emph{forward collision warning} (FCW), and \emph{intelligent speed advice} (ISA). These techniques share the common basic principle of detecting hazard scenarios, warning drivers of potential risks, and taking control of the vehicle in extreme situations. All these solutions require detection of driving anomalies that deviate from normal driving patterns, and increase chances of accidents. Current approaches for detecting abnormal driving behaviors or conditions often rely on either threshold-based or rule-based systems \cite{Hong_2014, Chakravarty_2013, Chen_2015_3, Dai_2010, Zhang_2017_5}. However, these methods are often triggered only when a driver makes a mistake, which is too late in many cases. Furthermore, it is highly unlikely that rule-based systems can exhaustively cover all potential anomaly scenarios. It is important to develop algorithms that can detect general abnormal driving behaviors as early as possible so that potential road accidents can be avoided. For these cases, unsupervised approaches are expected to be more effective, without rigid predefined rules or definitions of anomalies. 


This work proposes an unsupervised approach based on the conditional \emph{generative adversarial network} (GAN) framework to detect driving anomalies. Our approach is based on the premise that driving anomalies are often unexpected events that we cannot predict with the available contextual information. Motivated by this premise, we use the data from previous frames as a condition to generate prediction of signals of the near future from random noise. Then, we use the discriminator model to compare the differences between the generated prediction and real data of the upcoming frames, creating a powerful metric to  indicate the abnormal degree of the near future. This idea was previously validated in our preliminary work \cite{Qiu_2019_2}, where we implemented our approach with fully connected \emph{deep neural networks} (DNNs). In contrast, our proposed framework uses the latest advances in machine learning to leverage discriminative information directly from the data using \emph{convolutional neural networks} (CNNs). CNNs can extract both linear and non-linear relationships in and between sequences \cite{Borovykh_2017}, which we expect to be useful in detecting driving anomalies. Our formulation also leverages temporal information, relying on \emph{recurrent neural networks} (RNNs) implemented with \emph{long short-term memory} (LSTM) layers. LSTMs are designed to capture temporal relationship among sequential data from previous frames, providing a powerful framework for temporal sequence forecasting \cite{Fisher_2018}.

We build our approach with features extracted from the \emph{controller area network-bus} (CAN-Bus) and physiological signals, although this approach is flexible and can be implemented with different sensing modalities. CAN-Bus signals provide powerful information for estimating driving maneuvers and driver behaviors, including acceleration, breaks and steering wheel movements \cite{Zheng_2014,Sathyanarayana_2010,Jain_2011}. Therefore, we expect that predicting future CAN-Bus signals with our formulation will lead to robust driving anomaly detection. In addition to CAN-Bus data, we also rely on the driver's physiological signals. In particular, we consider \emph{Electrocardiography} (ECG), \emph{breath rate} (BR), and \emph{Electrodermal activity} (EDA) signals. The motivation for considering physiological signals is that under certain complex driving conditions, a driver might get nervous or frightened by abnormal driving events, which will be reflected in her/his biosignals. Even if the driver does not react with a driving maneuver, the physiological signals will indicate the presence of the anomaly. In fact, our preliminary study showed that adding features extracted from the driver's physiological data increased the model's discriminative power \cite{Qiu_2019_2}. Physiological signals are also closely related to driving behaviors \cite{Li_2016_3,Murphey_2015,Li_2016_4,Qiu_2019}. Consequently, our model considers the vehicle's CAN-Bus signals and driver's physiological signals.

We evaluate the proposed approach with the \emph{driving anomaly dataset} (DAD). This corpus has rich manual annotations of maneuvers and events. We group events that are likely to trigger driving anomalies, such as traffic violations, pedestrian on the road, and crossing vehicles. The anomaly scores of the driving segments overlapping with these annotations are generally higher than the anomaly scores of the driving segments without any annotations. We also investigate the contributions of the blocks in our formulation. For this purpose, we build our model with fully connected DNN, with only LSTMs, or with only CNNs. The result shows that the LSTM-based model performs better at discriminating abnormal from normal driving scenarios than the CNN-based model. The discriminative performance of our proposed approach is improved when we combine both structures (i.e., CNN and LSTM). These results are also confirmed with perceptual evaluations, where annotators were asked to assess the risk level and familiarity of video segments identified with high anomalous scores by the alternative models. The proposed CNN+LSTM based model is able to identify video segments that are perceived as more risky and less familiar by the annotators. In summary, the contributions of our study are:

\begin{itemize}[leftmargin=0.3em]
\vspace{-0.2em}
\setlength{\itemindent}{1em}
\setlength{\itemsep}{0cm}%
\setlength{\parskip}{0cm}%
\item We proposed an unsupervised formulation to predict driving anomalies based on conditional GAN, which contrasts predicted and observed physiological and CAN-Bus features. 

\item The proposed approach derives discriminative representations directly from data using CNNs, and leverages temporal information across consecutive frames using LSTMs. 

\item We validated the proposed approach with objective and perceptual evaluations using a naturalistic driving database, which demonstrates the strengths of our proposed formulation. 
\end{itemize}

The paper is organized as follows. Section \ref{sec:relatedWork} presents related studies, discussing effort to detect driving anomalies. It also discusses how conditional GANs have been used for anomaly/outlier detection in other fields. Section \ref{sec:data_base} introduces the dataset used in this study for evaluating our proposed models. Section \ref{sec:proposed_methods} presents the motivation of our framework, discussing the details of our formulation. Section \ref{sec:Experimental_results} evaluates the discriminative power of our unsupervised driving anomaly detection framework using objective and subjective evaluations. Finally, Section \ref{sec:conclusion} summarizes the contributions of this work, discussing potential ideas to extend and improve our unsupervised driving anomaly detection system.

\section{Related Work}
\label{sec:relatedWork}

This section reviews the most relevant studies to our work. We start with Section \ref{ssec:relatedDriving}, which presents studies aiming to detect driving anomalies. Section \ref{ssec:relatedGAN} presents a broader overview on anomaly detection methods using GAN across different areas. Section \ref{ssec:Prior_Work} highlights the differences between our work and other methods. 

\subsection{Driving Anomaly Detection}
\label{ssec:relatedDriving}

Pattern-based methods detect anomaly driving events by modeling driving maneuver patterns. Some of them identify cases where the driving behaviors depart from expected normal driving patterns, labeling them as abnormal events \cite{Mohan_2008, Aljaafreh_2012, Eren_2012, Nirmali_2017, Zhang_2017_5, Yang_2019, Ryan_202x}. Other studies have focused on detecting several specific types of abnormal driving maneuver patterns \cite{Dai_2010, Fazeen_2012, Xu_2014, Li_2016,Li_2013,Chen_2015_3, Ramyar_2016, Yu_2017, Chai_2017, El-Masry_2018}. Zhang \etal \cite{Zhang_2017_5} proposed a driving anomaly detection model by representing normal driving patterns in a state graph. They use this state graph as the criterion to distinguish abnormal driving behaviors that deviate from the expected state transitions. Another representative approach is the work of Chen \etal \cite{Chen_2015_3}. They considered six types of abnormal driving behaviors such as fast U-turn and sudden braking. They obtained the vehicle acceleration data from smartphone sensors, which were used to recognize these events using a \emph{support vector machine} (SVM). Dai \etal \cite{Dai_2010} detected driving under the influence of alcohol using a pattern-matching model, which compares the differences in the vehicle's acceleration between normal and drunk driving conditions.

The threshold based methods \cite{Mohamad_2011, Saiprasert_2013, Hong_2014, Chakravarty_2013, Wahlstrom_2014, Wahlstrom_2015, Li_2016_6, Liu_2016_2, Vavouranakis_2017} set bounds on the values of features or parameters describing the driving scenario. Abnormal driving conditions are set when their values are outside the predefined \emph{safe} ranges. Hong \etal \cite{Hong_2014} detected when the vehicle's acceleration was higher than a predefined threshold, using this event as a proxy to measure aggressive driving behaviors. Similarly, Chakravarty \etal \cite{Chakravarty_2013} proposed a system that evaluates multiple thresholds on the vehicle's acceleration to detect risky driving events. They considered the vehicle's acceleration on different directions to detect four maneuver types (i.e., hard bump, hard cornering, harsh brake, and sharp acceleration). Even though these methods are simple and computationally effective, the threshold-based methods using predefined values often lack flexibility, requiring, in many cases, domain knowledge (e.g., speed limit on current roads).

The clustering based method is another approach used to identify abnormal scenarios \cite{Constantinescu_2010, Zheng_2016, Hansen_2017, Hamdy_2017, Fugiglando_2019}. This approach builds on the hypothesis that most people drive in a proper and safe way under most naturalistic driving conditions. Therefore, abnormal driving behaviors or risky driving conditions are infrequent events. Under this assumption, 
we should expect that anomaly events will be clustered as outliers. Hansen \etal \cite{Hansen_2017} discriminated the driver's maneuvers by mapping the features extracted from the vehicle's dynamic signals to a feature space. The outlier driving events of the clusters were regarded as driving anomalies. The work of Zheng and Hansen \cite{Zheng_2016} used a one-class SVM and the \emph{topology anomaly detection} (TAD), clustering model to grade each driving event from ``good'' to ``bad''. The study established a four-class labeling framework, according to the clusters (from the innermost cluster to the outermost cluster).

\subsection{Anomaly Detection Using Conditional GANs}
\label{ssec:relatedGAN}

Multiple methods have been proposed for time series anomaly detection \cite{Chandola_2012}. An interesting formulation is the \emph{generative adversarial network} (GAN) \cite{Goodfellow_2014}, which has opened new directions for this problem. A GAN-based model consists of a generative model (\emph{G}) and a discriminative model (\emph{D}), which are trained with an adversary strategy. \emph{G} is trained to generate disruptive fake data from random noises, learning the distribution of the data that needs to be generated. As an adversarial game, \emph{D} is trained to identify the differences between the generated fake data and real data. As the quality of \emph{G} improves, the differences between the generated and real signals decrease, reducing the performance of \emph{D}. GAN has been widely used as a state-of-the-art generative model. This model has also been used for detection of anomalies or outliers. This section describe some examples of GAN-based anomaly detection models used in different domains. The reader are referred to Di Mattia \etal \cite{Di_Mattia_2019}, which presents a survey on this area. 

Schlegl \etal \cite{Schlegl_2017} presented the AnoGAN framework, which was applied to identify anomalies in retina tomography images. AnoGAN learns a manifold to represent the distribution of the data using a GAN. It maps the image into a latent space where it is feasible to quantify deviations from normal distributions, detecting anomalous cases. Zhou \etal \cite{Zhou_2019_2} proposed BeatGAN, an unsupervised GANs-based system to identify unusual human motions (e.g., jumping and running) from normal motions (i.e., walking). This approach built a generator with an encoder-decoder structure, using the reconstructed signals as the fake signals to confuse the discriminator. For the evaluation, they used the reconstruction error between the real signal and the generated fake signal as the anomaly metric to detect abnormal motions. Xue \etal \cite{Xue_2018} proposed a supervised approach based on GAN, called SegAN, for brain tumor segmentation. The generator creates segmentation masks for the input brain image. The discriminator identifies between generated segmentations and ground truth labels. The formulation regards the tumor area as the abnormal part of a given image. Li \etal \cite{Li_2018_3} proposed a GAN-based model to detect attacks on \emph{cyber-physical systems} (CPSs). The approach compares the predictions of the generator with actual multivariate time-series data. The residual between these signals is used to detect anomaly activities in the CPS. Studies have also used GAN for \emph{out-of-domain} (OOD) detection in \emph{natural language understanding} (NLU). For example, Zheng \etal \cite{Zheng_2020} trained an autoencoder to map the input utterance into the latent embedding created by the generator of a GAN model. Then, they used the decoder of the autoencoder to generate utterance from the embedding as OOD samples. 

Rather than generating data merely from random noise, Mirza \etal \cite{Mirza_2014} modified the original GAN formulation by adding extra information as condition to the model. The additional input conditions the data generation process, creating behaviors that are properly constrained. Hyland \etal \cite{Hyland_2017} adopted the conditional GAN framework implemented with LSTMs to generate fake patients' \emph{heart rate} (HR) and \emph{respiratory rate} (RR) data, conditioning on blood pressure values. They used this method to detect patients' abnormal physical conditions. Akcay \etal \cite{Akcay_2018} proposed the GANomaly framework, which is another conditional adversary network for anomaly detection. The approach combines GAN with an autoencoder to jointly model a latent and image space. The encoder processes the input data creating a latent space. The decoder processes the latent space to create a reconstructed version of the data. The discriminator aims to classify the original image as real and the reconstructed image as fake. Then, the encoder is used again to map the reconstructed image back into the latent space. The distance in the latent space between the input data and reconstructed latent vector is used as the anomaly score. The model is trained with neutral data. Anomalous examples that do not fit the normal distribution are expected to have larger distance. A similar approach was used by Zenati \etal \cite{Zenati_2018} in their \emph{efficient GAN based anomaly detection} (EGBAD) framework, without using the encoder on the reconstructed image. 


\subsection{Relation to Prior Work}
\label{ssec:Prior_Work}

Our proposed approach builds upon our preliminary work in Qiu \etal \cite{Qiu_2019_2}. This study extracted statistic features from the data to capture the key aspects of the signals as the input of the model. The approach has two problems that are addressed in this study. First, the model used hand crafted features from the modalities.  Our proposed architecture addresses this problem by using the CNN block, which extracts discriminative representations directly from the data. The second limitation is that the model only considers static windows of six seconds to constraint the model. This approach ignores temporal information within these previous six second, since statistics are derived from the entire segment. It also ignores longer dependencies that may be important in the prediction of the signals in the upcoming frames. The proposed approach addresses this limitation by integrating the LSTM module. With the combination of both additions, we extract discriminative temporal representations directly from the raw data. We believe that this feature representation can reveal more detailed information than statistic features using predefined functional module. Qiu \etal \cite{Qiu_2020} used an architecture that was similar to the work proposed in Qiu \etal \cite{Qiu_2019_2}, but with an anomaly score relying on the triplet loss function. We do not explore this direction in this paper. We provide an exhaustive evaluation of our proposed architecture, comparing the benefits of adding the CNN and LSTM blocks. We also compares the proposed approach with several alternative methods.

While other studies have proposed GAN-based approaches for anomaly detection, the particular formulation presented in this study is novel, and has important benefits with respect to other alternative models. Our model learns from real data and make forecasts of the upcoming signals based on the observed data. We use the discriminator to determine segments with signals that deviate from the observed patterns. Our approach is fully unsupervised, where we only need to collect data without the need for labels. Most of the other GAN-based methods are supervised (e.g., SegAN \cite{Xue_2018}). GAN-based anomaly detection methods, such as AnoGAN \cite{Schlegl_2017} and GANomaly \cite{Akcay_2018} are trained merely with normal data, and tested with normal/abnormal data. They detect anomalies by discriminating the samples that are different from the normal ones, calling these samples as \emph{abnormal}. Our approach is fundamentally different from these formulations, where we train the model with all the data using a predictive formulation. It is also important to notice that all these studies using GAN-based approaches for anomaly detection were designed for other problems in other fields. With the exception of our preliminary study \cite{Qiu_2019_2,Qiu_2020}, this is the first time that conditional GAN has been used for driving anomaly detection.

\section{Driving Anomaly Dataset (DAD)}
\label{sec:data_base}

This work uses the \emph{driving anomaly dataset} (DAD). The corpus includes 250 hours of naturalistic driving recordings in an Asian city collected by the Honda Research Institute in collaboration with a local company. One experienced driver participated in the data collection process driving a Honda Accord. This dataset includes the driver's physiological data, which is a key modality in this study. The driver's \emph{electrocardiography} (ECG) and \emph{breath rate} (BR) signals are recorded with a Zephyer BioHarness 3 chestband. The ECG signal is recorded at 250 Hz, and the BR signal is recorded at 25 Hz. The data collection also included the \emph{Electrodermal activity} (EDA) signals from the driver obtained with a Empatica E4 wristband collected at 4 Hz. To compensate for the differences across sessions, we normalize the physiology data per session by using the Z-normalization. The corpus also provides the vehicle's CAN-Bus data. We obtain six vehicle signals: speed, steering speed, steering angle, throttle angle, brake pressure, and yaw. These signals are recorded at 100 Hz. All the vehicle's CAN-Bus data and driver's physiological data are synchronized at 30 Hz.

The data collection also includes data from other sensors not used in this study. The setup includes three FLIR Blackfly S cameras facing the road [(i.e., right, center and left)]. The data collection also included Tobii Pro 2 eye-tracking glasses. 

\begin{table}[t]
\caption{Annotations included in the DAD database.}
\begin{center}
\begin{tabular}{p{1.8cm} | p{6cm}}
\hline
 & Annotations  \\
\hline
Goal-oriented Operation & Left turn; Right turn; Intersection passing; Crosswalk passing; Left lane change; Right lane change; U-turn\\
\hline
Stimulus-driven Operation & Stop for congestion; Avoid pedestrian near ego lane; Avoid road motorcyclist; Avoid on-road bicyclist\\
\hline
Traffic rule$/$manner violation & Traffic rule violation\\
\hline
Attention & Crossing vehicle; Crossing pedestrian; Red light; Cut-in; Sign; On-road bicyclist; Parked vehicle; Merging vehicle; Yellow light; Road work; Pedestrian near ego lane\\
\hline
\end{tabular}
\end{center}
\label{tab:annotation}
\end{table}

\begin{figure*}[t]
  \centering
  \includegraphics[width=\linewidth]{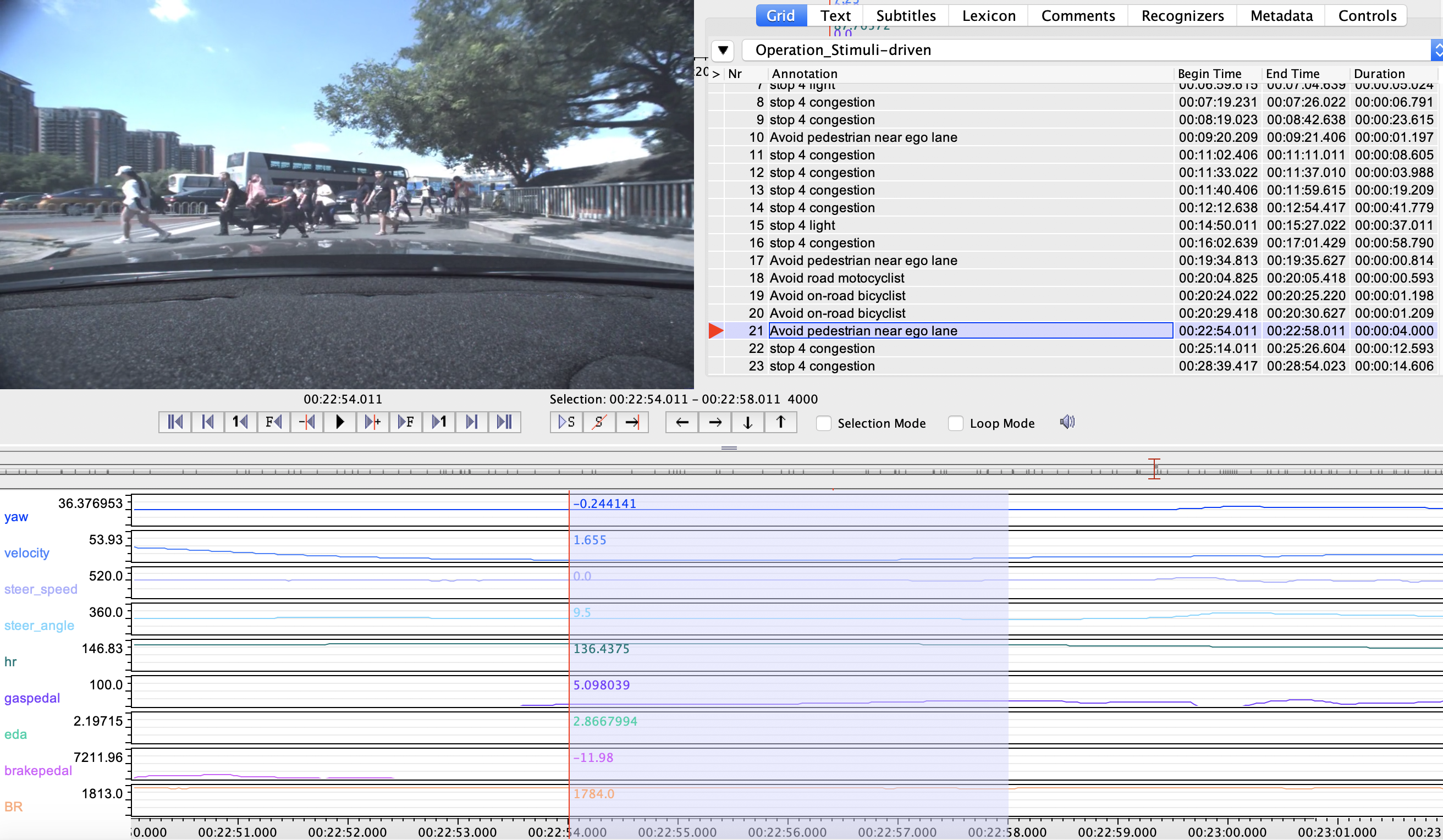}
  \caption{The software ELAN showing some of the annotations included in the DAD database. The annotations, physiological data, and CAN-Bus data are synchronized with the road videos.}
  \label{fig:ELAN}
\end{figure*}

One of the strengths of the DAD corpus is the rich set of annotations, which follows the approach used in the collection of the \emph{Honda Research Institute driving dataset} (HDD) \cite{Misu_2018,Ramanishka_2018}. The annotations are grouped into four layers: goal-oriented operation, stimuli-driven operation, traffic rule/manner violation, and driver's attention. Table \ref{tab:annotation} shows the specific annotations within each layer (see study of Ramanishka \etal \cite{Ramanishka_2018} for details on this annotation process). The annotations of the driver's driving maneuvers are manually added to the dataset. The annotations and the sensing modalities are aligned, including the drivers' physiological signals and the vehicles' CAN-Bus data. For clearer visualization, all the data sequences, annotations, and videos are combined and synchronized using the open source software ELAN. Figure \ref{fig:ELAN} shows the \emph{user interface} (UI) of ELAN, where the driving events correspond to the \emph{stimulus-driven operation} layer.

While the corpus includes 250 hours of data, not all the recordings have been annotated. We only use 121 sessions, which include about 130 hours of well-annotated urban driving recordings. We split these recordings into train (100 sessions, $\sim$105 hours), development (11 sessions, $\sim$13 hours), and test (10 sessions, $\sim$12 hours) sets. The DAD corpus does not have annotations for anomaly scores. Instead, we evaluate our model using the existing annotations overlapping with the driving videos used to test our framework. For this purpose, we group the driving events in the test set into three groups. The first group is the  \emph{candidate} set, which consists of traffic rule violations and hazard driving conditions such as avoiding on-road pedestrians or parked vehicles. We expect that these events will include segments with high driving anomaly scores. The second group is the \emph{maneuver} set, which includes segments annotated with regular driving maneuvers such as right turns, intersection passing, and U-turns. In general, we expect moderate anomaly scores for these events. The third group is the \emph{normal} set, which includes segments without any annotation. We expect low anomaly scores for these segments. Table \ref{tab:sets} lists the events associated with each group. Notice that our approach is fully unsupervised, so the annotations are exclusively used to evaluate our approach.

\begin{table}[t]
\caption{Sets of annotations considered to evaluate the proposed anomaly detection models. The candidate set is expected to have driving scenarios that can be considered as anomalous.}
\vspace{-0.2cm}
\begin{center}
\begin{tabular}{p{1.8cm} | p{6cm}}
\hline
Sets  & Annotations  \\
\hline
Candidate & Avoid on-road pedestrian; Avoid pedestrian near ego-lane; Avoid on-road bicyclist; Avoid bicyclist near ego-lane; Avoid on-road motorcyclist; Avoid parked vehicle; traffic rule violation\\
\hline
Maneuver & Left turn; Right turn; Left lane branch; Right lane branch; U-turn; Intersection passing\\
\hline
Normal & No annotations during the segments\\
\hline
\end{tabular}
\end{center}
\label{tab:sets}
\end{table}

\section{Proposed Conditional GAN Model}
\label{sec:proposed_methods}

The goal of this study is to implement an unsupervised framework to detect driving anomalies. A driving anomaly is something unexpected that deviates from normal patterns. We are not just interested on detecting dangerous conditions. Instead, this work focuses on unexpected driving events. An ADAS should be able to leverage knowledge of unexpected events, even if they do not represent dangerous scenarios. Dangerous events are special cases, since, in daily life, some dangerous driving scenarios are usually caused by unexpected maneuvers or reactions from traffic participants (e.g., drivers of other vehicles or the ego-vehicle, pedestrians, bicyclists and motorcycle). Therefore, we expect that risky and hazardous scenarios will be considered by our system as driving anomalies. While the framework is general and can be implemented with different features, we detect abnormal driving behaviors using the vehicle CAN-Bus data and the driver's physiological data. The CAN-Bus data consists of the vehicle's speed, yaw, pedal angle, brake pressure, steer angle, and steering speed. The physiological data includes HR, BR, and EDA signals.  Our motivation for using physiological signals is that HR, BR, and EDA respond to mental and cognitive states, indicating stress \cite{Rompelman_1977}, and anxiety \cite{Taelman_2009} levels. Previous studies on driver behaviors have used physiological data \cite{Healey_2005,Nishigaki_2018}, showing correlation with driving maneuvers \cite{Qiu_2019,Murphey_2015, Li_2016_3,Li_2016_4}. These findings show that physiological signals can be natural complements to CAN-Bus signals, providing information even when a driver fails to maneuver the car in the presence of unexpected events.

\begin{figure}[t]
  \centering
  \includegraphics[width=0.97\columnwidth]{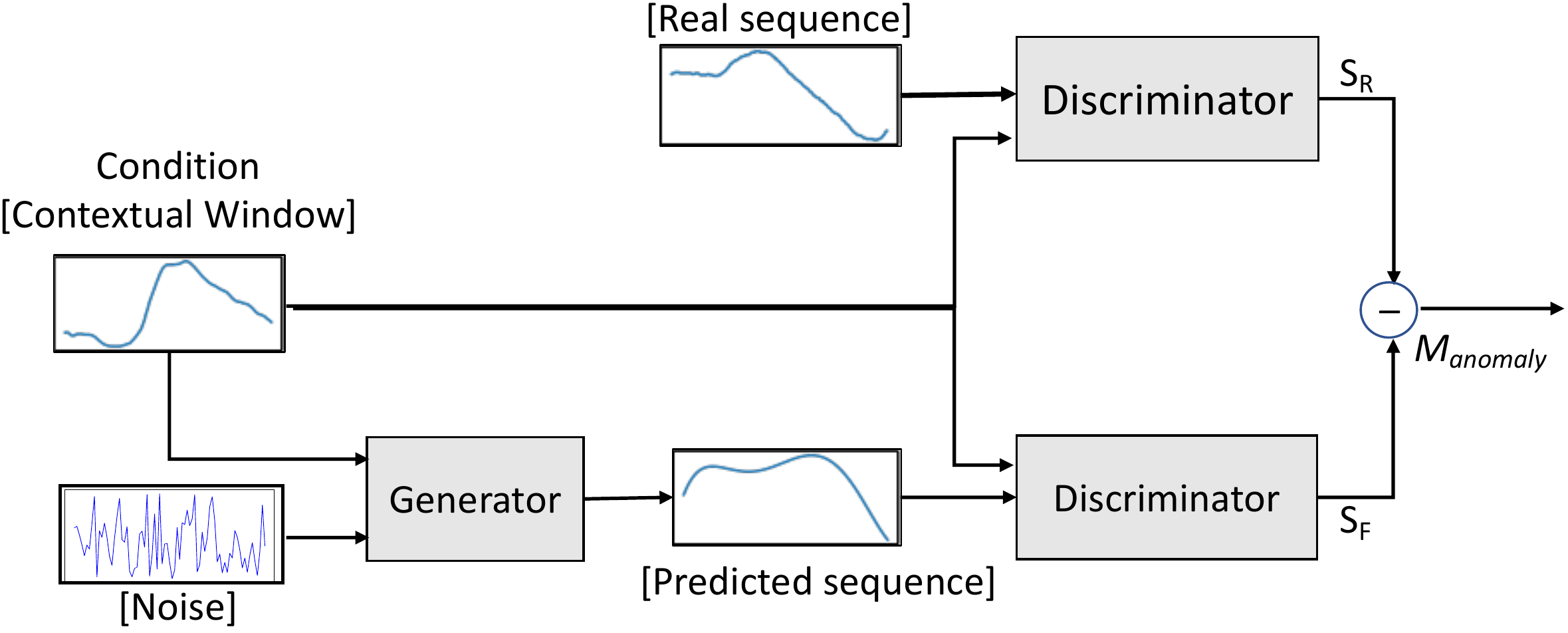}
  \caption{Abstract illustration of our GAN-based model for driving anomaly detection. Conditioned by a contextual window with previous frames, the generator predicts the features in the near future. The discriminator takes the predictions and the real data as inputs comparing the value of the discriminator's score. Unexpected events are then identified with this unsupervised model.}
  \label{fig:generating_procedure}
\end{figure}

The premise of our method is that anomaly scenarios are often associated with unexpected events. Therefore, we predict the features from upcoming driving events, conditioned on the previous values of the target features. Then, we quantify the difference between the predicted driving data and the actual data. We implement these ideas with conditional GAN, which is one of the most powerful generative models.

Figure \ref{fig:generating_procedure} shows the procedure of our implementation. The generator creates plausible data sequences conditioned on the previous values of the features. The discriminator recognizes whether the input data is real or fake (i.e., created by the generator). The scores from the discriminator are used to determine our anomaly scores. This section describes our proposed approach in detail. First, we present the intuition of our formulation (Sec. \ref{ssec:CGAN}). Then, we present a basic implementation with fully connected layers (Sec. \ref{ssec:CGAN-fc}), where we explain the main features of our formulation. Sections \ref{ssec:CGAN_CNN} and \ref{ssec:lstm_based_CGAN} introduce the use of CNN to extract discriminative features directly from data, and LSTM to capture temporal information. Finally, Section \ref{ssec:cnn_lstm_based_CGAN} presents our full model, which combines the architectures of the LSTM and CNN based models.

\subsection{Anomaly Detection with Conditional GAN}
\label{ssec:CGAN}

Multiple studies have revealed the impressive capability of GANs to learn the distribution of target data, generating similar samples from random noise. This learning procedure is accomplished through an adversarial game between the discriminator and generator, as shown in Equations \ref{equ:Discriminator} and \ref{equ:Generator}.


\begin{equation} 
\label{equ:Discriminator}
\begin{split}
\underset{D}{\max}V(D) =& \mathbb{E}_{\boldsymbol{x} \sim p_{data} (\boldsymbol{x})}\left[ \log D(\boldsymbol{x}) \right] \\
&+  \mathbb{E}_{\boldsymbol{z} \sim p_{z} (\boldsymbol{z})}\left[ \log (1 - D(G(\boldsymbol{z}))) \right]
\end{split}
\end{equation}

\begin{equation}
\underset{D}{\min}V(D) = \mathbb{E}_{\boldsymbol{z} \sim p_{z} (\boldsymbol{z})}\left[ \log (1 - D(G(\boldsymbol{z})) ) \right]
\label{equ:Generator}
\end{equation}

The discriminator outputs a value $D(\boldsymbol{x})$ which indicates whether the input $\boldsymbol{x}$, with probability distribution $p_{data}$, is a real sample. The objective of $D(\boldsymbol{x})$ is to maximize the chance to identify the real sample as real, and the generated fake samples as fake. The discriminative score is a sigmoid output which ranges from 0 to 1, where 1 means absolutely real and 0 means absolutely fake. The objective function of the generator aims to create the prediction $G(\boldsymbol{z})$ as close as possible to real samples to fool the discriminator. The variable $\boldsymbol{z}\sim p_z$ is a noise vector used as the input. Following  Equations \ref{equ:Discriminator} and \ref{equ:Generator}, the GAN model should be trained to converge to a good estimator of $p_{data}$. While the input of a regular GAN is the random noise vector  $\boldsymbol{z}$, a conditional GAN uses extra information ($\boldsymbol{y}$) as additional input to constrain the model to generate more targeted predictions. In our study, the condition of the generative model is the data from previous frames. Figure \ref{fig:generating_procedure} describes the training process of the proposed conditional GANs, showing that the input of \emph{G} is random noise and the data sequence from previous frames as condition. The output is the generated data sequence of the upcoming analysis window, $G(\boldsymbol{z|y})$. 


Figure \ref{fig:generating_procedure} shows the inference process of the proposed model. Given the analysis window, the generator creates a fake data sequence. Then, \emph{D} takes either the real data from the upcoming analysis window or the fake data generated by \emph{G}. For each of these inputs, \emph{D} creates a score ($S_{R}$ for real signal; $S_{F}$ for fake/generated signal). The difference between the scores is regarded as the anomaly score, $m_{anomaly}$ (Eq. \ref{eq:metric}). This metric represents the uncertainty of the upcoming driving events, which we hypothesize to be an informative driving anomalous metric. For a well-trained generative model, \emph{G} generates realistic data from noise in order to confuse \emph{D}. Therefore, when the real data from the analysis window follows the distribution of the regular data, the generated signal will be similar to the real data and $S_{F}$ will be closer to $S_{R}$.  This case will produce a small value for $m_{anomaly}$, indicating that the input of \emph{G} is normal and predictable.  However, if the upcoming driving data in the analysis window has a distribution that differs from the expected behaviors, the data will be unpredictable, creating a gap between $S_{F}$ and $S_{G}$. This scenario will have a larger $m_{anomaly}$ value. A key advantage of our approach is that it only requires unlabeled data, providing an appealing unsupervised formulation.

\begin{equation}
m_{anomaly} = \left |  S_{R} - S_{F}\right |
\label{eq:metric}
\end{equation}

\subsection{Conditional GAN with Fully Connected Layers}
\label{ssec:CGAN-fc}

In Qiu \etal \cite{Qiu_2019_2}, we presented a preliminary implementation of our unsupervised driving anomaly detection, where the discriminator and generator were implemented with \emph{fully connected} (FC) layers. The approach uses a fixed window with previous values of physiological and CAN-Bus data to predict future values of the data. The discriminator takes the statistical features of either the real or generated signals as input. The feature set includes four time domain features from each of the CAN-Bus data and physiological data (i.e., maximum, minimum, mean, and standard deviation). From each physiological data, we additionally extracted five frequency domain features, calculating the energy in the frequency domain covering the following five bands: [0-0.04 Hz], [0.04-0.15 Hz], [0.15-0.5 Hz], [0.5-4 Hz], and [4-20 Hz]. The generator is implemented with five layers, each of them implemented with 180 neurons. Similarly, the discriminator is implemented with five layers, each of them with 51 neurons. 

\subsection{CNN-based Conditional GAN}
\label{ssec:CGAN_CNN}

The first improvement for our model is replacing the feature extraction module. Instead of using predefined functionals over the previous frames, we learn directly discriminative patterns from the data using CNNs. Models designed based on CNNs have been successful for end-to-end classification tasks \cite{Krizhevsky_2012}. The study of Borovykh \etal \cite{Borovykh_2017} showed that CNNs are able to extract temporal features from time series data that were discriminative to predict upcoming data values. Inspired by these studies, we implement our conditional GAN models with CNNs to learn more discriminative features directly from data. 

\begin{figure}[t]
  \centering
  \subfigure[Generator of the CNN-based model]{
  	\includegraphics[width=0.97\columnwidth]{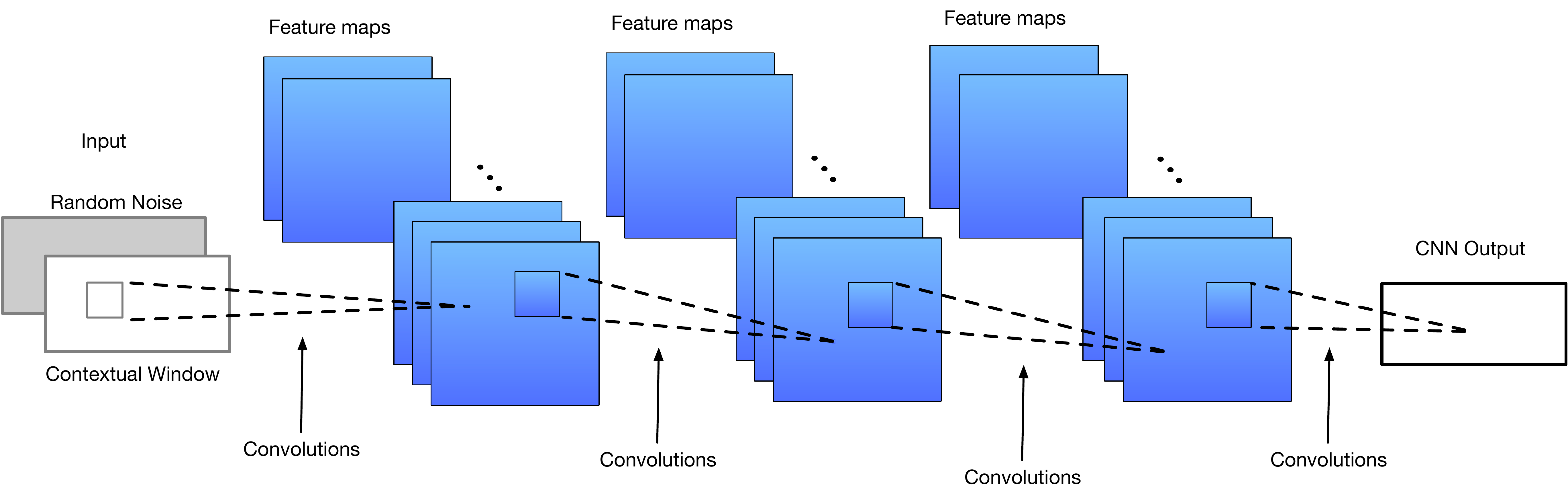}
	\label{fig:CNN_G}
  }\\
  \subfigure[Discriminator of the CNN-based model]{
  	\includegraphics[width=0.97\columnwidth]{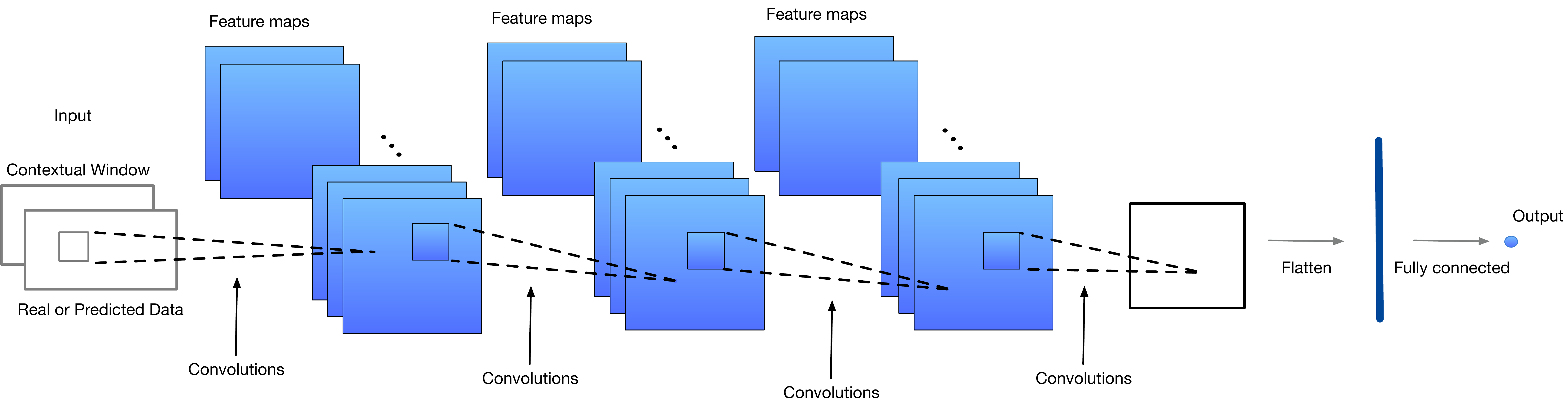}
	\label{fig:CNN_D}
  }
  \caption{Implementation of proposed CNN-based GAN model. The generator and discriminator have similar architectures with four CNNs, extracting discriminative representations directly from physiological and CAN-Bus data.}
  \label{fig:CNN_GAN}
\end{figure}

Figure \ref{fig:CNN_GAN} shows the implementation of our CNN-based GAN model. The generator and the discriminator consist of four convolutional layers, implemented with 18, 18, 9, and 1 channels, respectively. The kernel size for each layer is 9, 3, 3, and 3, respectively. We add a fully connected layer after the convolutional layers. During the training process, \emph{D} and \emph{G} are trained for 20 epochs and the Adam learning rate is set to 0.001. From our previous work \cite{Qiu_2019}, we conclude that features extracted from CAN-Bus and driver's physiological data with an analysis window of 12 seconds can be used to classify driving maneuvers. We aim to reduce the size of the window to reduce the latency in the model. However, the analysis window cannot be reduced too much to capture changes on physiological signal. As a compromise, we set the analysis window size to six seconds for our CNN-based GAN model. The model takes as input the previous six-second data and random noise, and generates predictions for the upcoming six-second data describing the upcoming CAN-Bus and physiological signals.

\subsection{LSTM-based Conditional GAN}
\label{ssec:lstm_based_CGAN}

A limitation of the CNN-based GAN model is that the contextual information is limited to fixed size windows (i.e., six seconds). We hypothesize that modeling longer temporal relationships in the data can lead to better results. We explore this direction by using RNNs, which can extract temporal relationships in the data. We implement the RNNs with \emph{long short-term memory} (LSTM) cells, leveraging the longer history information from previous time series data. Sequence to sequence tasks can be effectively implemented with RNN-based model, as demonstrated in language translation \cite{Sutskever_2014}. Additionally, models based on conditional RNNs have been successfully applied on tasks that mimic a particular writer's handwriting style \cite{Graves_2013_2}. Borovykh \etal \cite{Borovykh_2017} demonstrated the reliable capability of LSTM to make short-term prediction on trends in the stock market. Motivated by these studies, we design our approach with LSTM cells, expecting to improve the temporal modeling of our framework while avoiding rigid contextual feature vectors.

\begin{figure}[t]
  \centering
  \subfigure[LSTM-based Generator]{
  	\includegraphics[width=0.46\columnwidth]{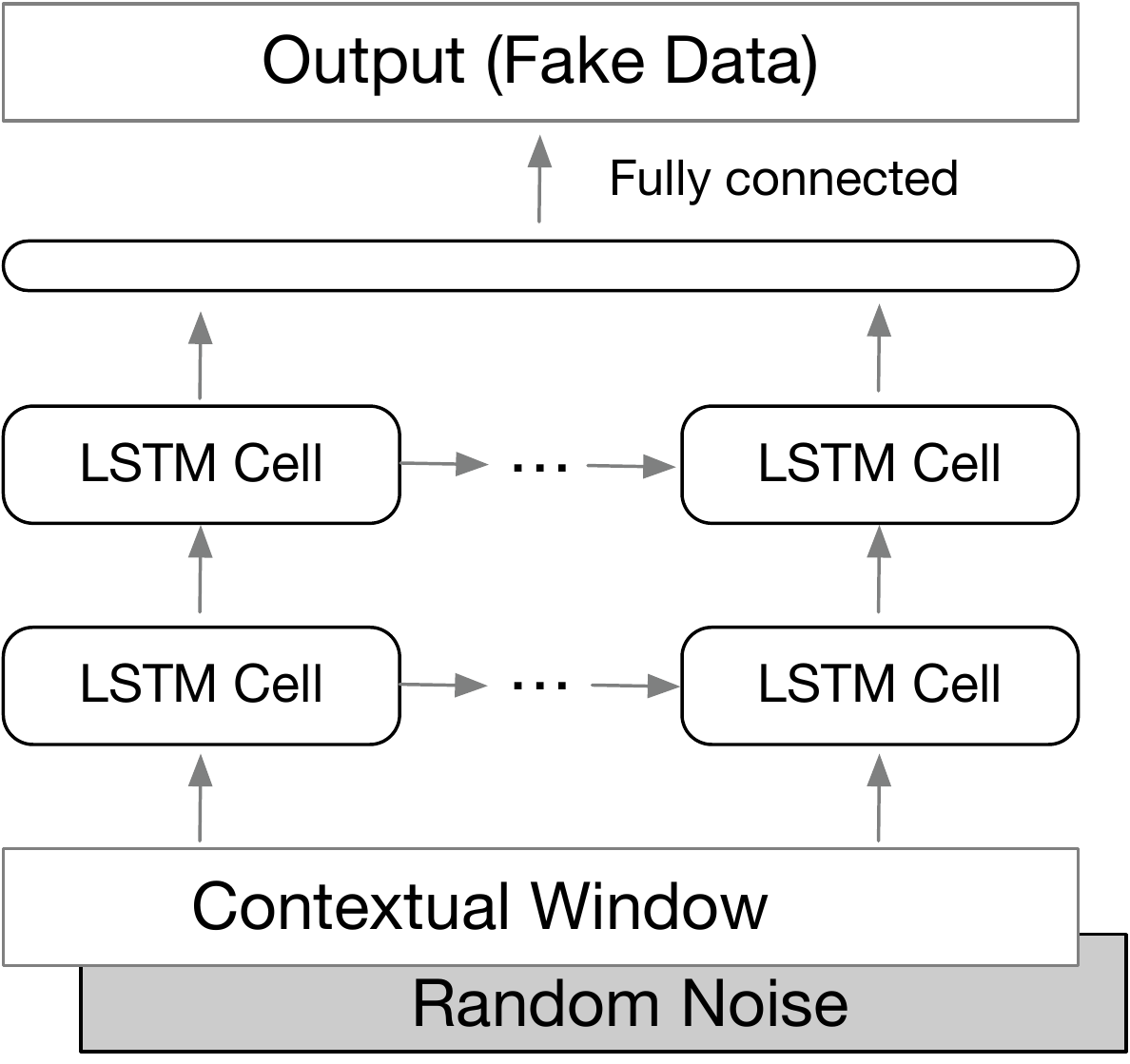}
	\label{fig:LSTM_G}
  }
  \subfigure[LSTM-based Discriminator]{
  	\includegraphics[width=0.46\columnwidth]{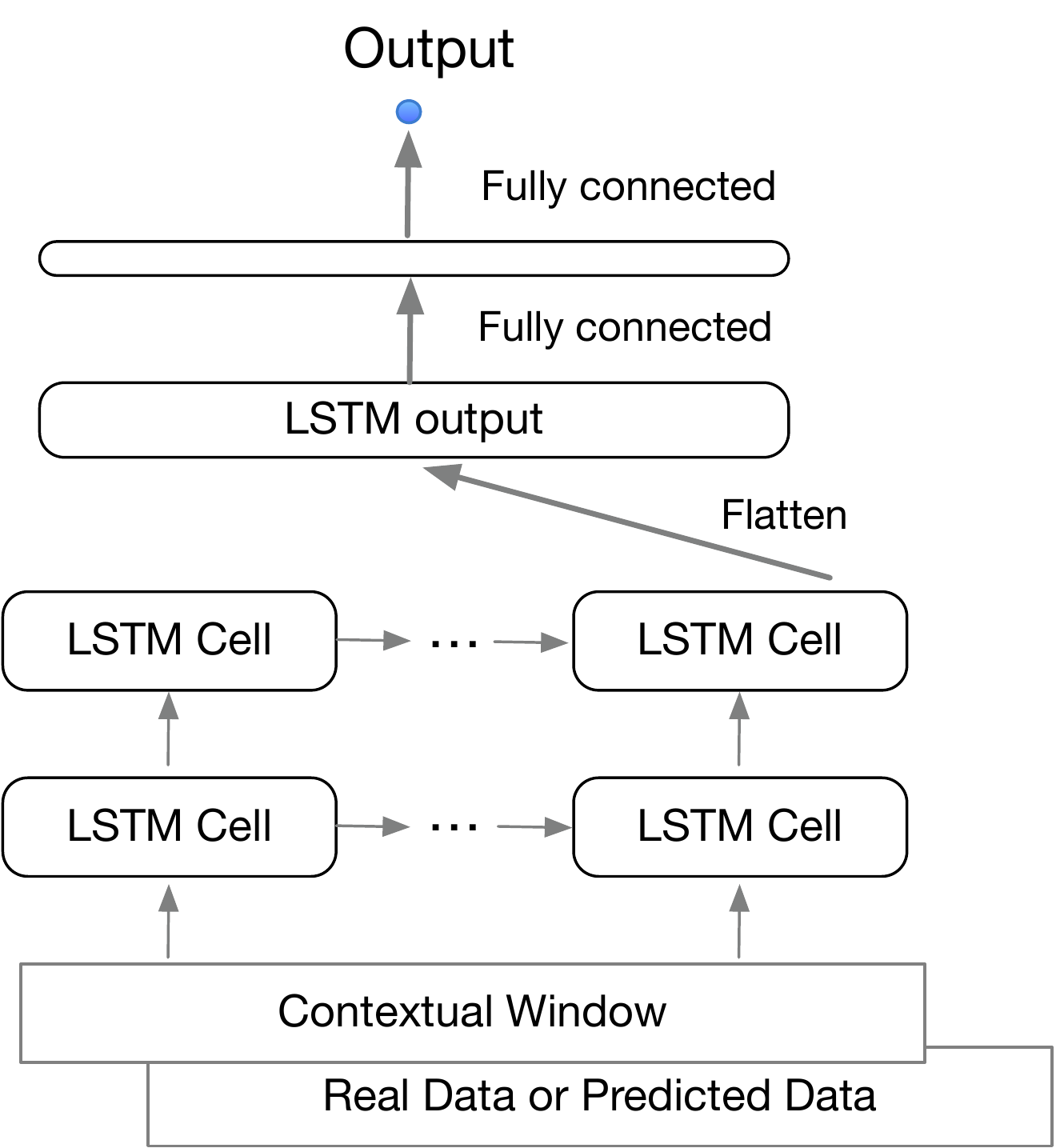}
	\label{fig:LSTM_D}
  }
  \caption{Implementation of the LSTM-based GAN model. The model relies on longer contextual window to leverage short and longer dependencies in the data to detect driving anomalies.}
  \label{fig:LSTM_GAN}
\end{figure}

Figure \ref{fig:LSTM_GAN} shows the structure of our LSTM-based GAN model. The model consists of two layers of LSTM cells, each of them implemented with 27 nodes. The generator of the LSTM-based GAN model takes a longer contextual window than the window analysis in the CNN-based GAN model, relying on the last 60 seconds of data. The generator predicts the next six seconds of physiological and CAN-Bus signals. The extended contextual window allows the LSTM cells to more effectively leverage short and long temporal relationships to make the predictions. The structure of the discriminator is similar to the generator, as shown in Figure \ref{fig:LSTM_D}. The discriminator takes six-second data, which can be real signals or data predicted by the generator, conditioned on the contextual window with the previous 60-second sequence. We extract the last output of the LSTM cells, flattening the feature representation as a vector. We add a fully connected layer creating a one-dimensional output, which predicts if the data is real or predicted by the generator. We use this score to estimate the anomaly score in Equation \ref{eq:metric}.


\begin{figure}[t]
  \centering
  \subfigure[Generator of the CNN+LSTM-based model]{
  	\includegraphics[width=0.97\columnwidth]{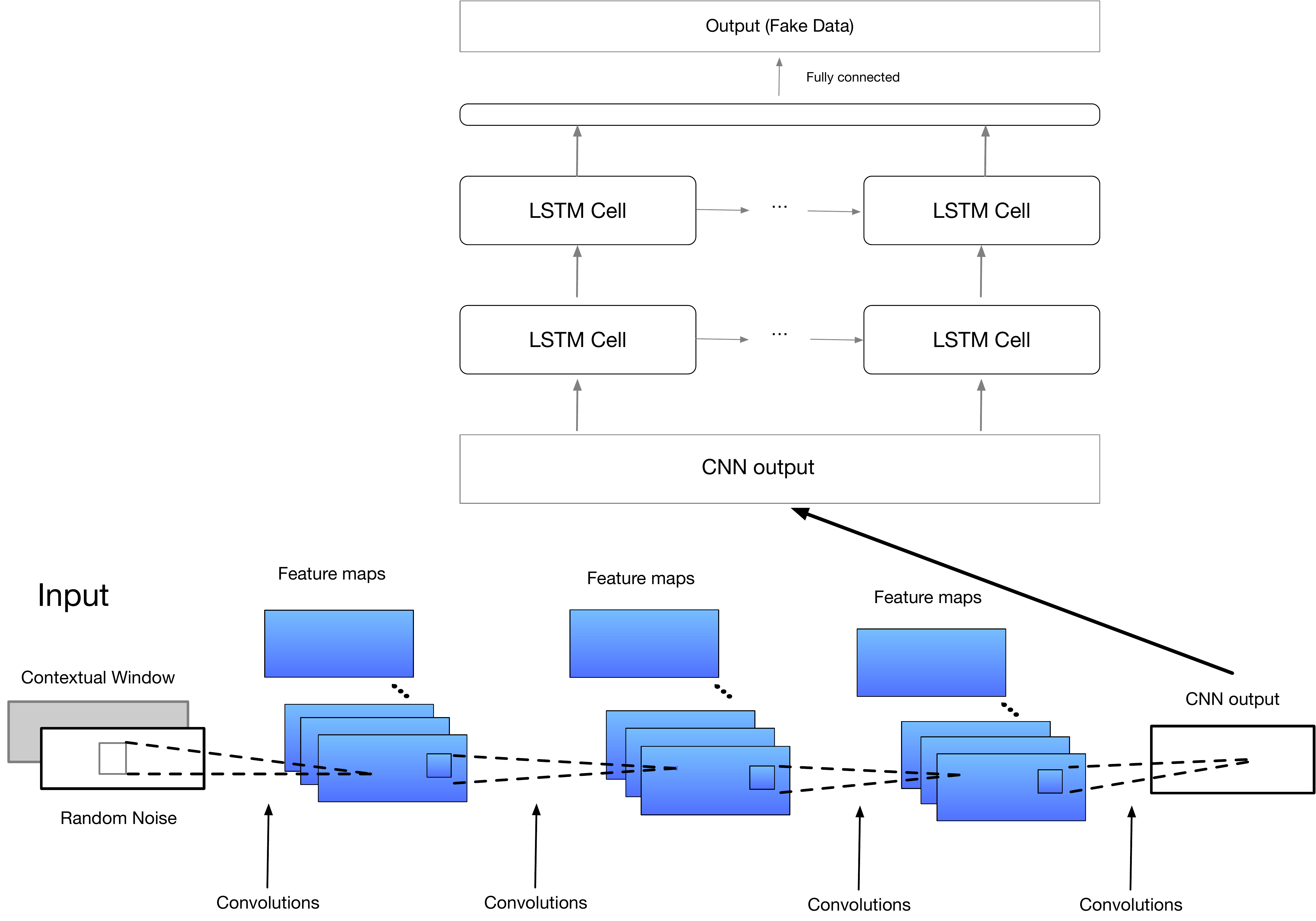}
	\label{fig:CNN_LSTM_G}
  }\\
  \subfigure[Discriminator of the CNN+LSTM-based model]{
  	\includegraphics[width=0.97\columnwidth]{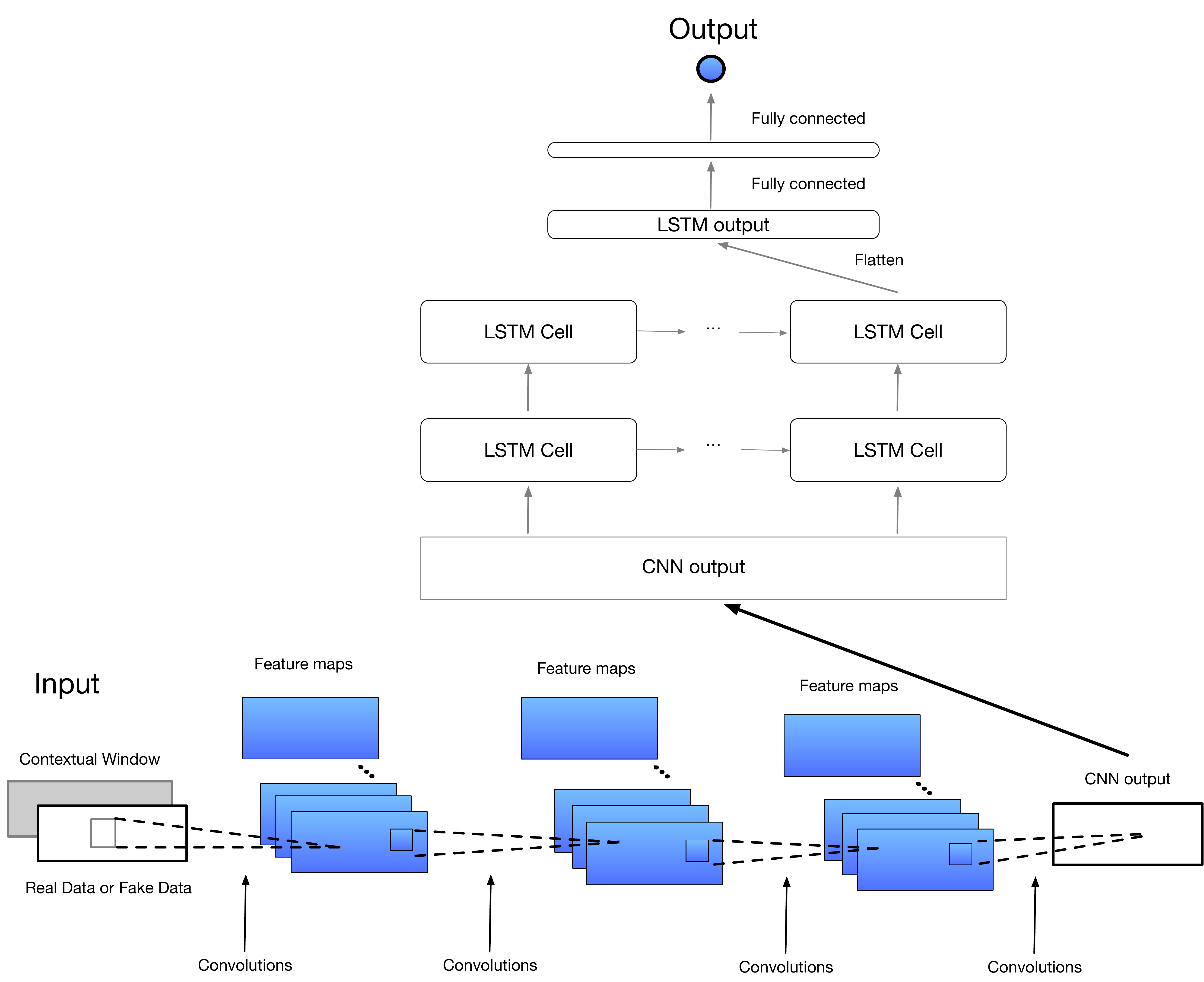}
	\label{fig:CNN_LSTM_D}
  }
  \caption{Implementation of the proposed CNN+LSTM-based GAN model. The unsupervised approach combines the strengths in using CNNs and LSTMs, extracting discriminative representations directly from the data while leveraging temporal information.}
  \label{fig:CNN_LSTM_GAN}
\end{figure}

\subsection{CNN+LSTM-Based Approach using Conditional GAN}
\label{ssec:cnn_lstm_based_CGAN}

The CNN-based and LSTM-based GAN models offer complementary benefits for our task. Therefore, our final model combines their structures leveraging better feature representations and temporal modeling. Figure \ref{fig:CNN_LSTM_GAN} shows the implementation of the proposed CNN+LSTM-based conditional GAN model, where we use the same structures presented for the CNN-based GAN model (Sec. \ref{ssec:CGAN_CNN}) and the LSTM-based GAN model (Sec. \ref{ssec:lstm_based_CGAN}) as blocks to build this model. First, the CNN block of the generator extracts discriminative information from the random noise and contextual window with the previous sixty-second used to condition the models. We implement the CNN block by splitting the contextual window into 10 six-second segments, without overlap. Then, we concatenate the CNN output of each of the six-second segments, creating a conditional embedding, which is used as the input of the LSTM block. The output of the LSTM is the prediction of the physiological and CAN-Bus data for the next 6-second window. For the discriminator, we implement the model using the same structure used for the generator. The only difference is the output layer, which is a one-dimensional score to predict whether the data is real or fake.

During the training process, we first import the pre-trained parameters of the CNN-based and LSTM-based GAN models. Then, we train the LSTM parameters (including both the generator and the discriminator) for 10 epochs while freezing the CNN parameters. Then, we jointly train the entire model together for another 10 epochs to get the final model.

\section{Experimental Results}
\label{sec:Experimental_results}
\begin{figure}[t]
  \centering
  \subfigure[Generator and discriminator losses in the training set]{
    \includegraphics[width=6.5cm]{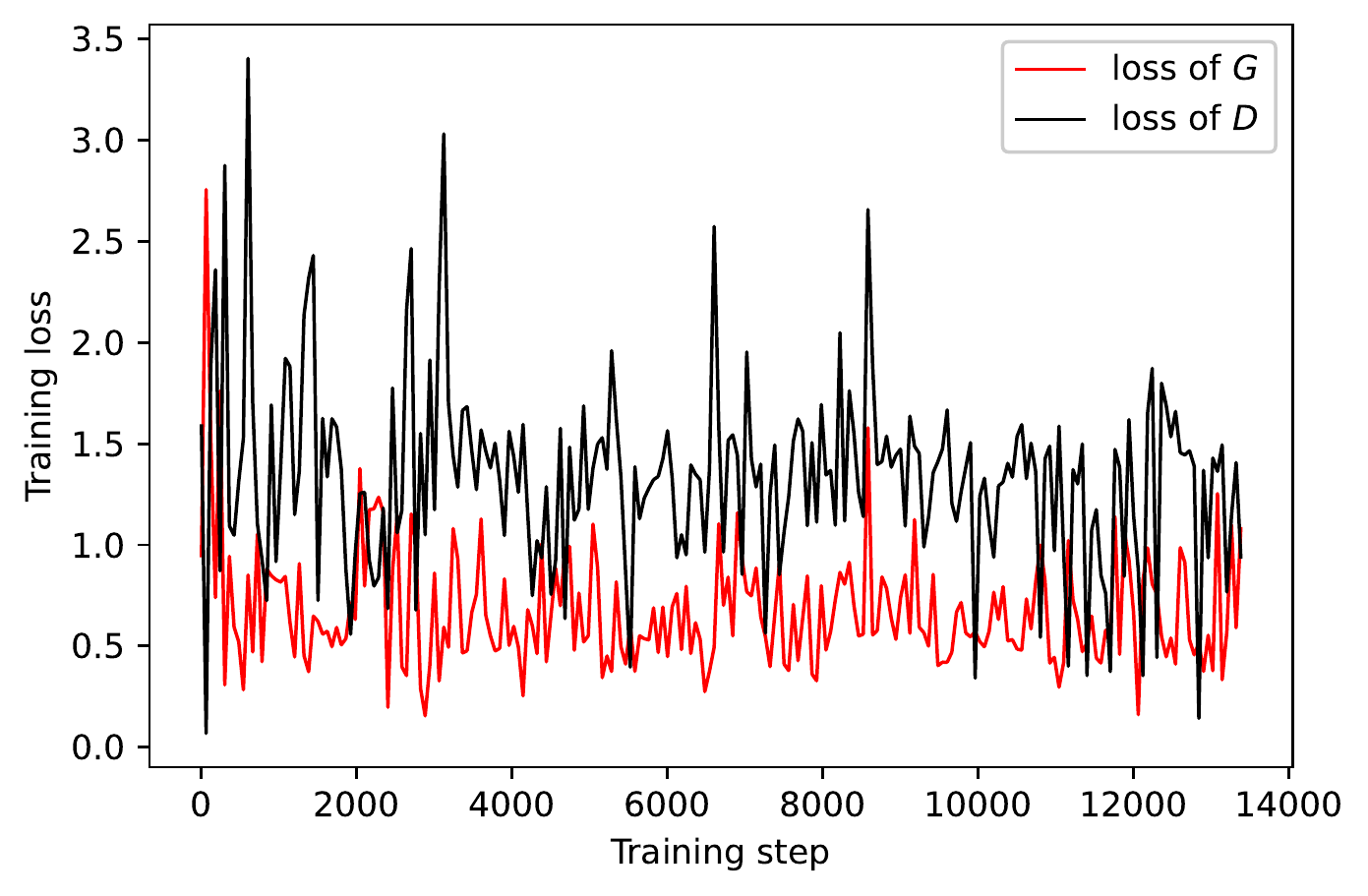}
    \label{fig:training_loss_score_DG}
  }
  \subfigure[Discriminator performance on the development set]{
    \includegraphics[width=6.5cm]{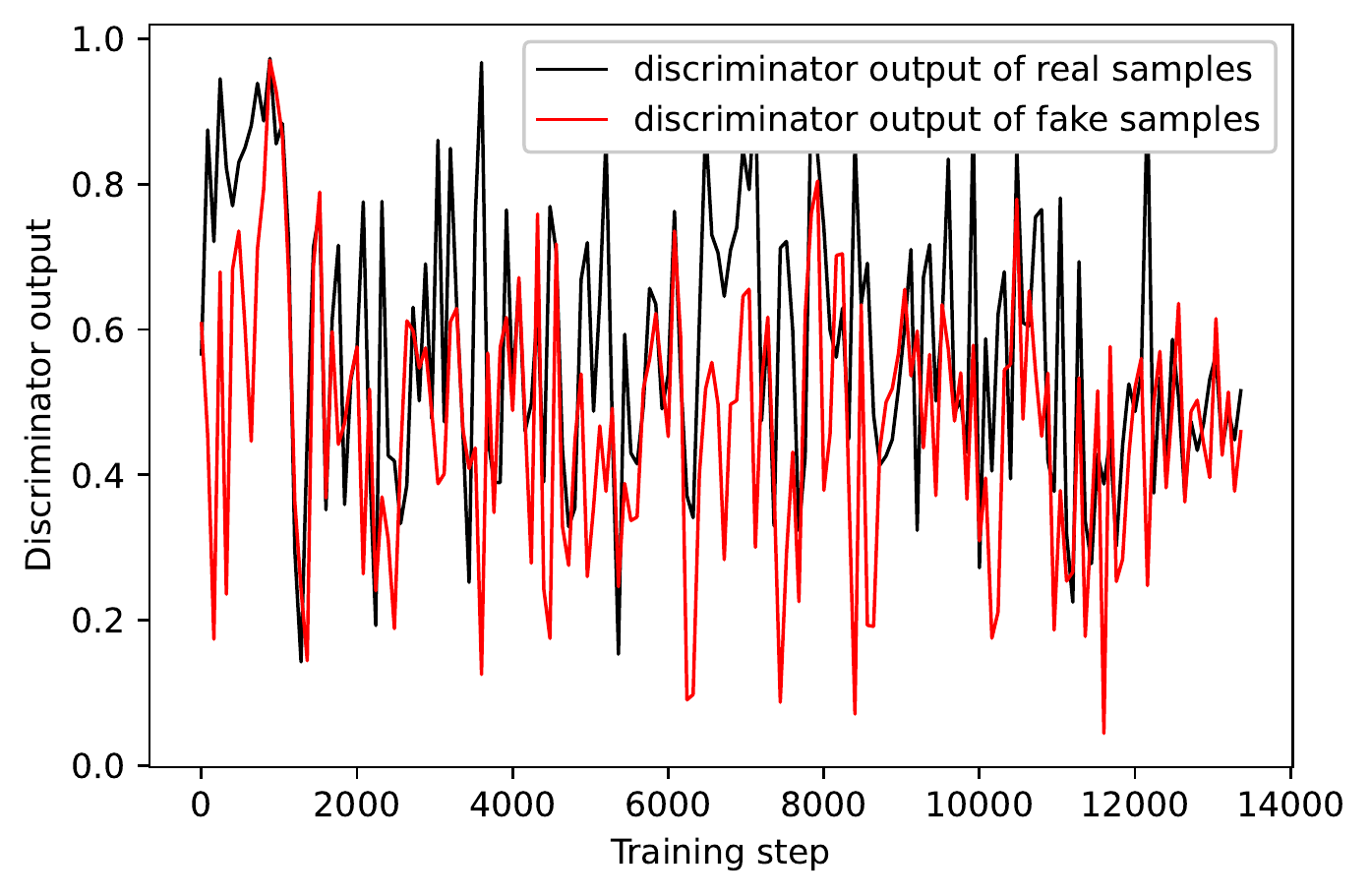}
    \label{fig:training_loss_score_D}
 }
  \caption{Loss of the GAN network during training. The training losses of \emph{D} and \emph{G} are often constantly fluctuating, as both losses go down. The outputs of the discriminator for the fake and real signals oscillate around 50\%.}
  \label{fig:training_loss_score}
\end{figure}

This section describes the experimental results obtained with our proposed conditional GAN models. Figure \ref{fig:training_loss_score_DG} shows the losses of the generator and discriminator in the training set. Training a GAN is not always easy, since it is a minmax optimization process. It is a problem if the loss of the discriminator drops fast compared to the generator's loss. A strong discriminator means that it is easy to distinguish between fake and real samples. Without an appropriate feedback from the discriminator, it is hard to train the generator. When properly trained, the overall loss of the GAN is often constantly fluctuating, as both losses go down. This is the exact pattern observed in Figure \ref{fig:training_loss_score_DG}. For a properly trained GAN, the discriminator should have problems recognizing between real and fake samples. The classification performance should be 50\% if the number of real or generated samples are equal. Figure \ref{fig:training_loss_score_D} shows the probability of the output of the discriminator for real and generated samples on the development set (1 is real, 0 is fake). The performance oscillates around 50\% for both type of samples, as expected. These figures shows that the GAN is properly trained.

\begin{figure}[t]
\centering
 \includegraphics[width=0.95\columnwidth]{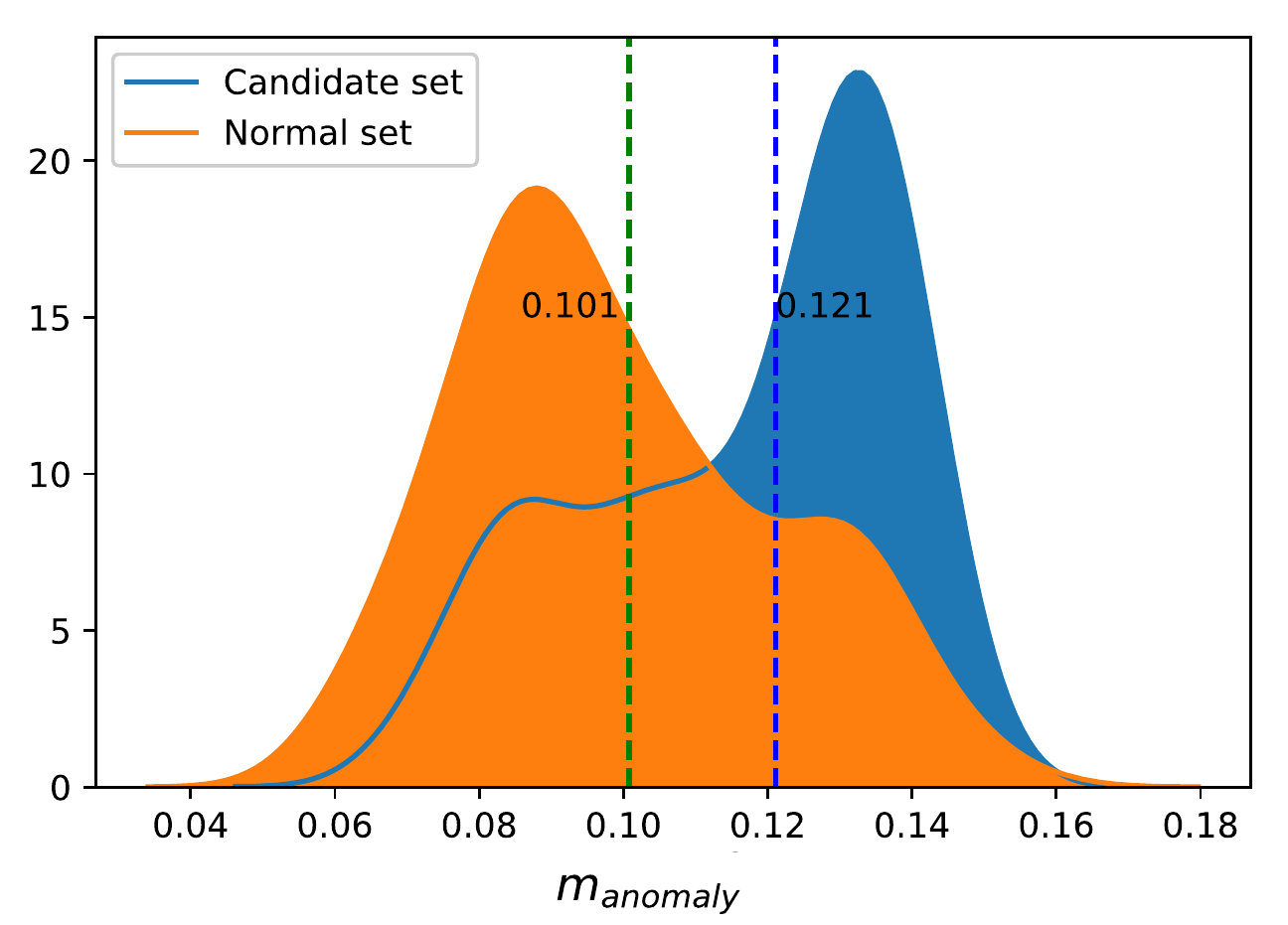}
  \caption{Histogram of the predictions of the FC-based models for segments in the test set from the normal and candidate sets. The vertical dashed lines indicate the medians of the anomaly scores for the sets.}
  \label{fig:FC_hist}
\end{figure}
We evaluate the performance by comparing the scores obtained from videos in the \emph{candidate}, \emph{maneuver}, and \emph{normal} sets. We also evaluate the performance with perceptual evaluations.

\subsection{Distribution of Anomaly Scores}
\label{subsec:score_distribution}

As stated in Section \ref{sec:data_base}, driving events in the \emph{candidate} set are more hazardous and rarely seen than the events in the \emph{normal} set. Therefore, the anomaly scores ($m_{anomaly}$) of the driving events in the \emph{candidate} set are expected to be larger than the corresponding scores on the \emph{normal} set. The analysis in this study compares the distribution of anomaly scores of the driving events from these two sets. For each segment, we provide the previous data as condition using six seconds for the FC-based and CNN-based models, and 60 seconds for models implemented with LSTMs. All models generate predictions for the upcoming six seconds.

\begin{figure*}[t]
\centering
\subfigure[CNN -- Physiological Signal]{
 \includegraphics[width=5.5cm]{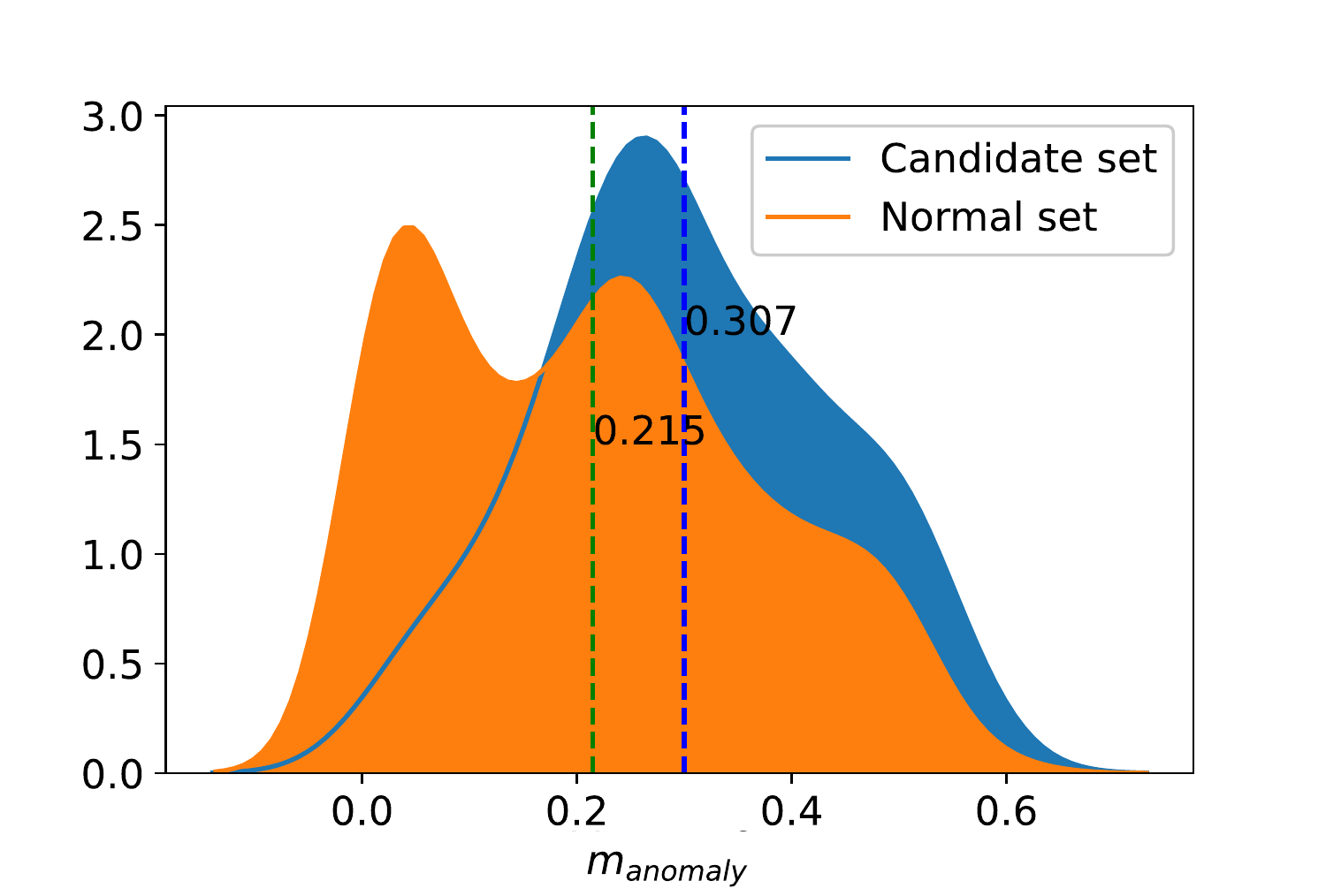}
  \label{fig:CNN_Physiological}
  }
\subfigure[CNN -- CAN-Bus Signal]{
 \includegraphics[width=5.5cm]{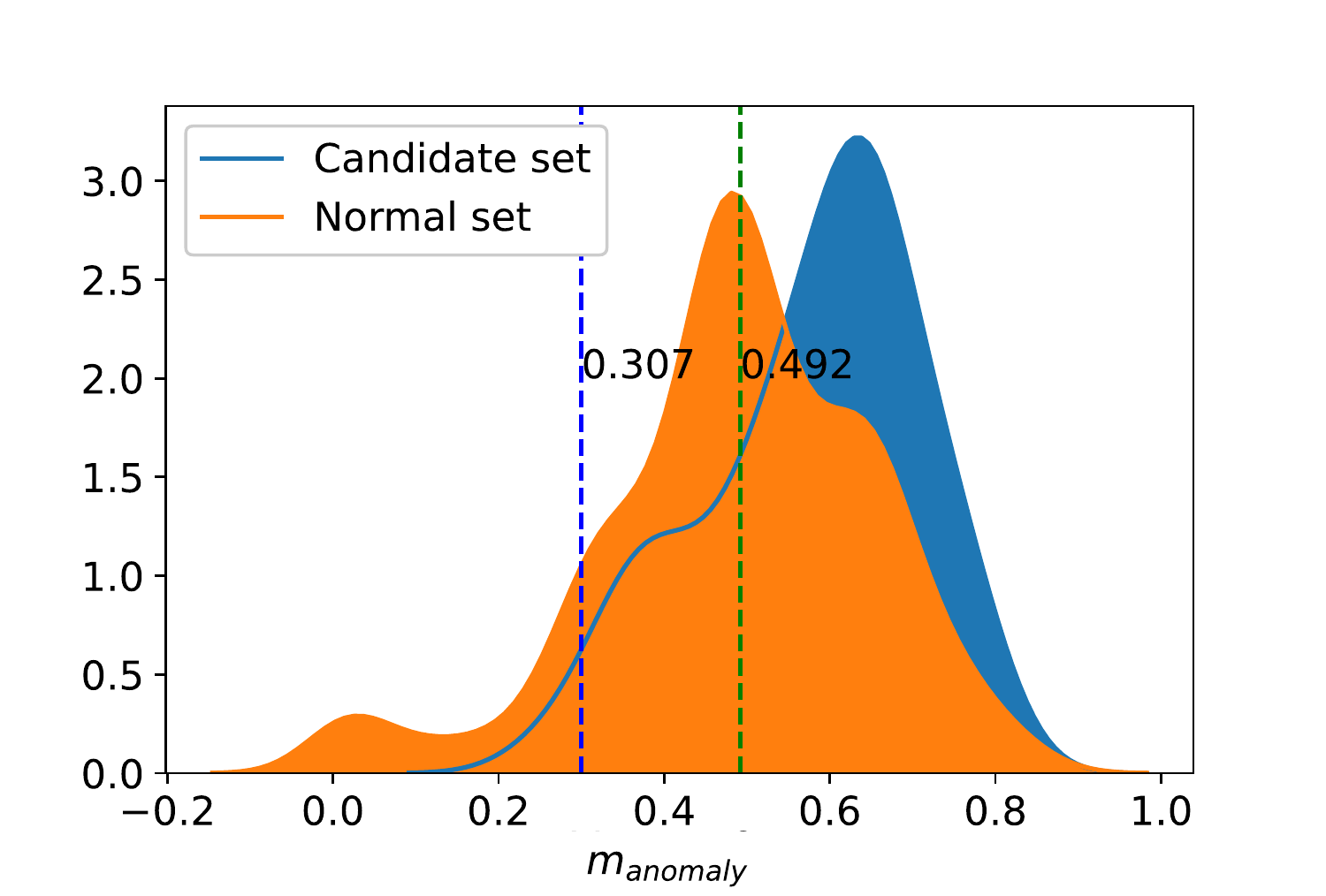}
  \label{fig:CNN_CANBUS}
  }
\subfigure[CNN -- All combined]{
 \includegraphics[width=5.5cm]{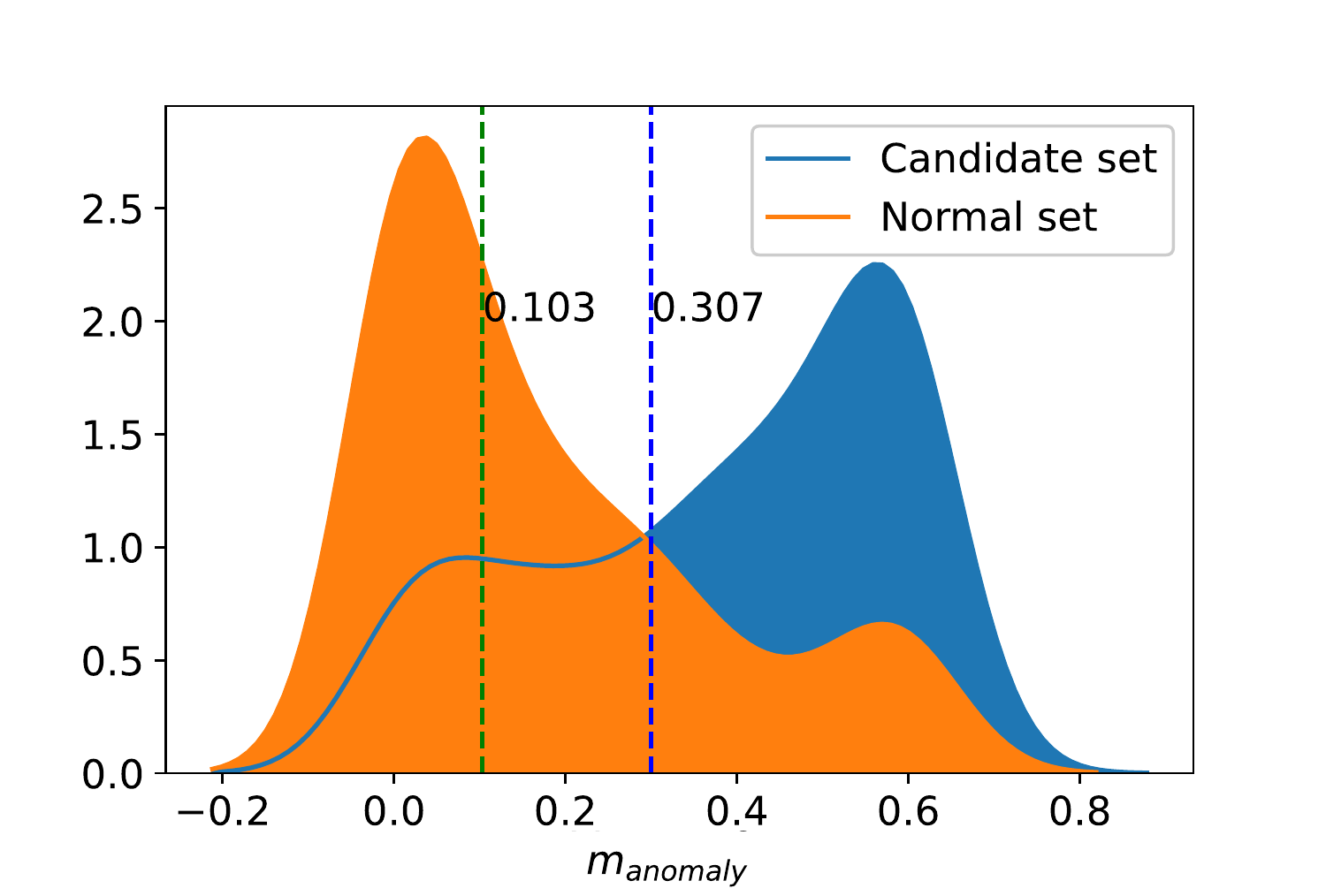}
  \label{fig:CNN_ALL}
  }\\
  \subfigure[LSTM -- Physiological Signal]{
   \includegraphics[width=5.5cm]{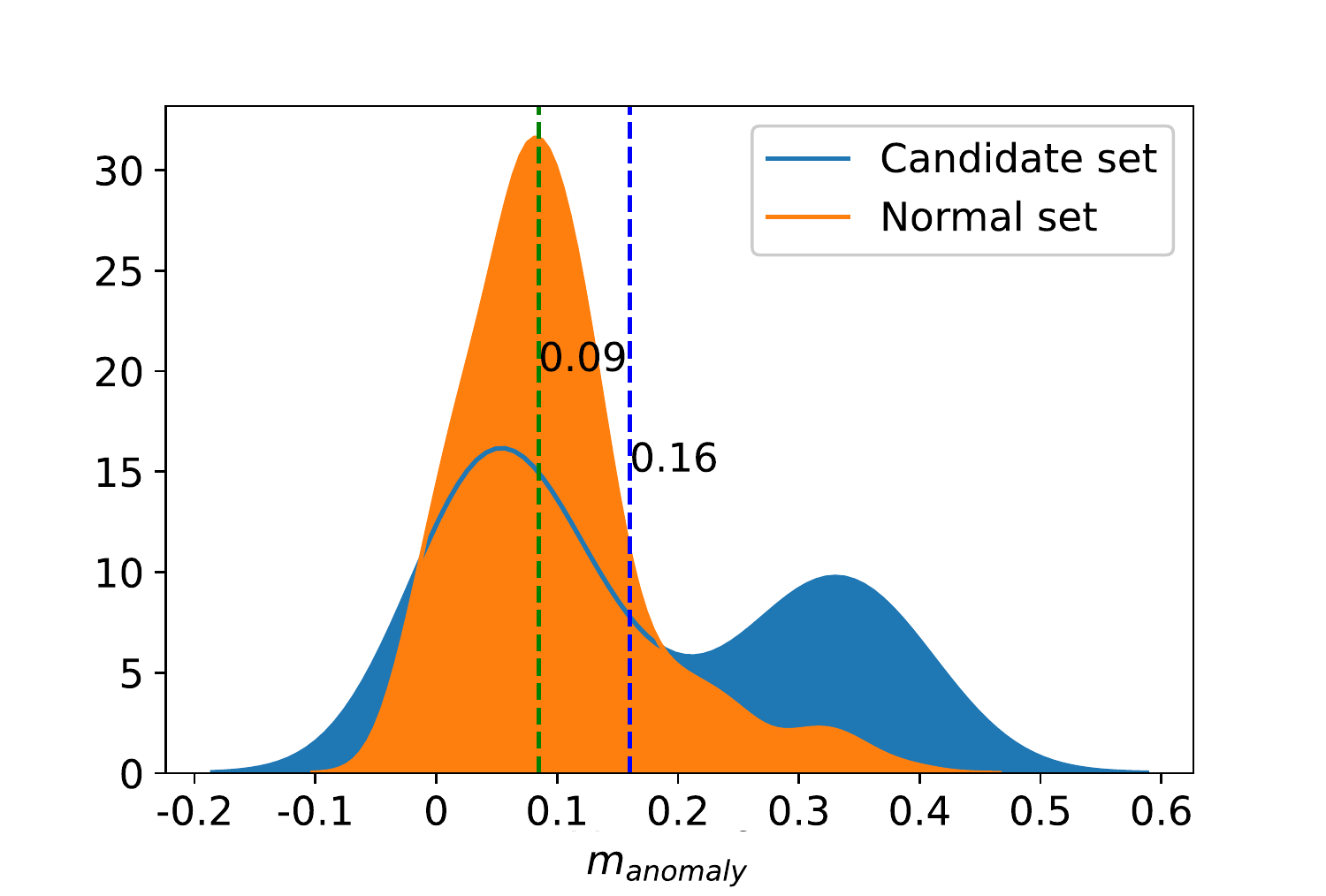}
    \label{fig:LSTM_Physiological}
    }
  \subfigure[LSTM -- CAN-Bus Signal]{
   \includegraphics[width=5.5cm]{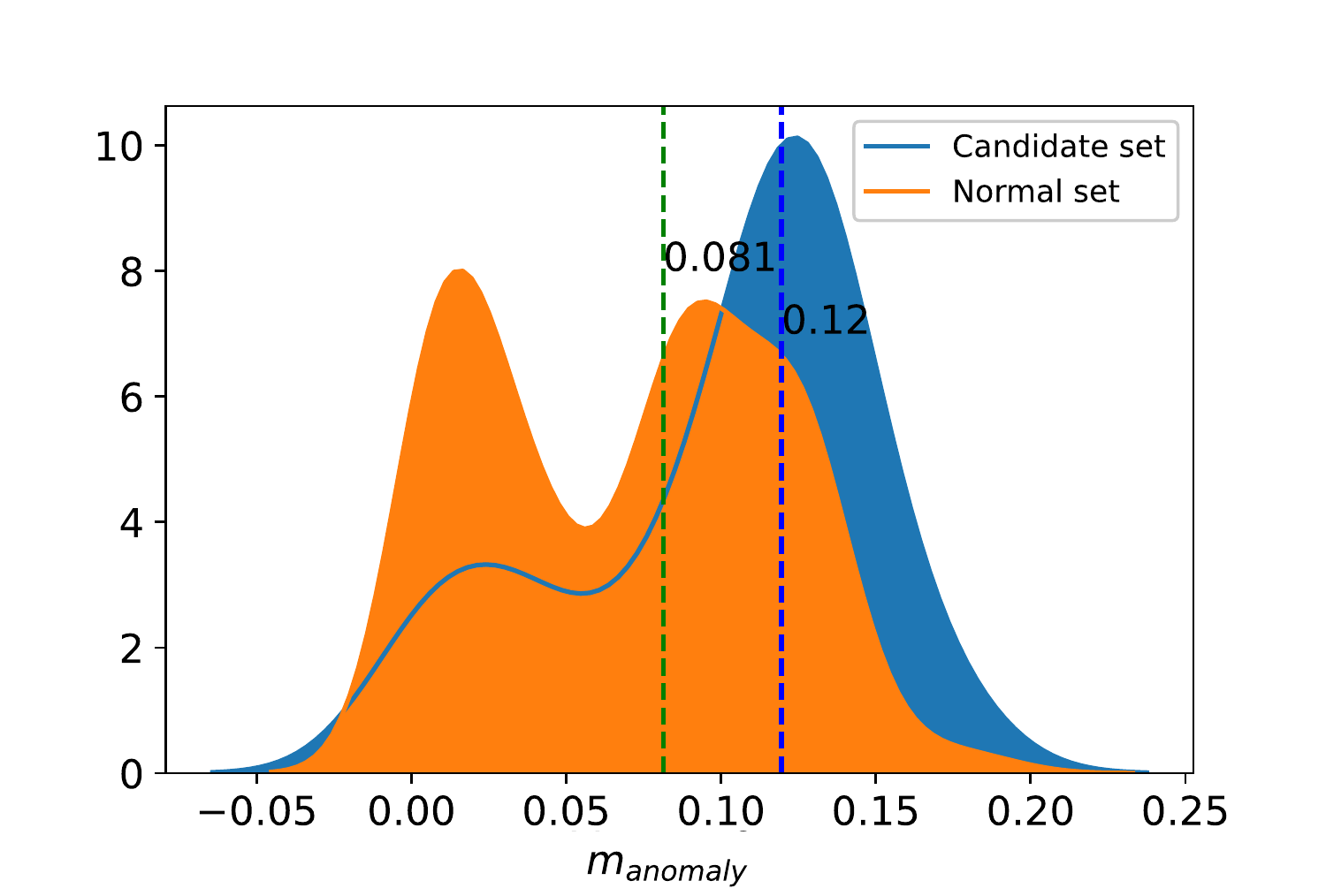}
    \label{fig:LSTM_CANBUS}
    }
  \subfigure[LSTM -- All combined]{
   \includegraphics[width=5.5cm]{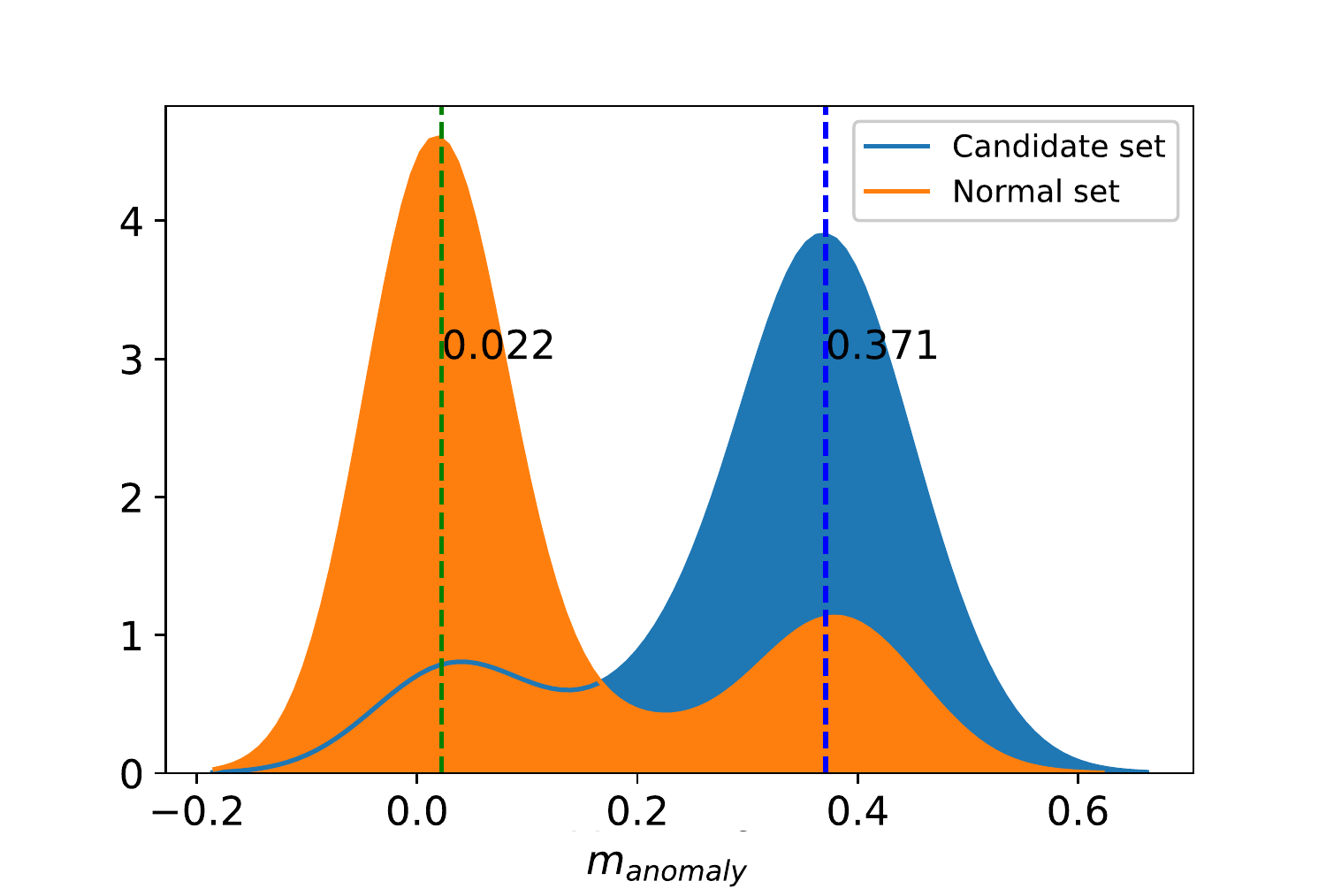}
    \label{fig:LSTM_ALL}
    }
    \\
    \subfigure[CNN+LSTM -- Physiological Signal]{
     \includegraphics[width=5.5cm]{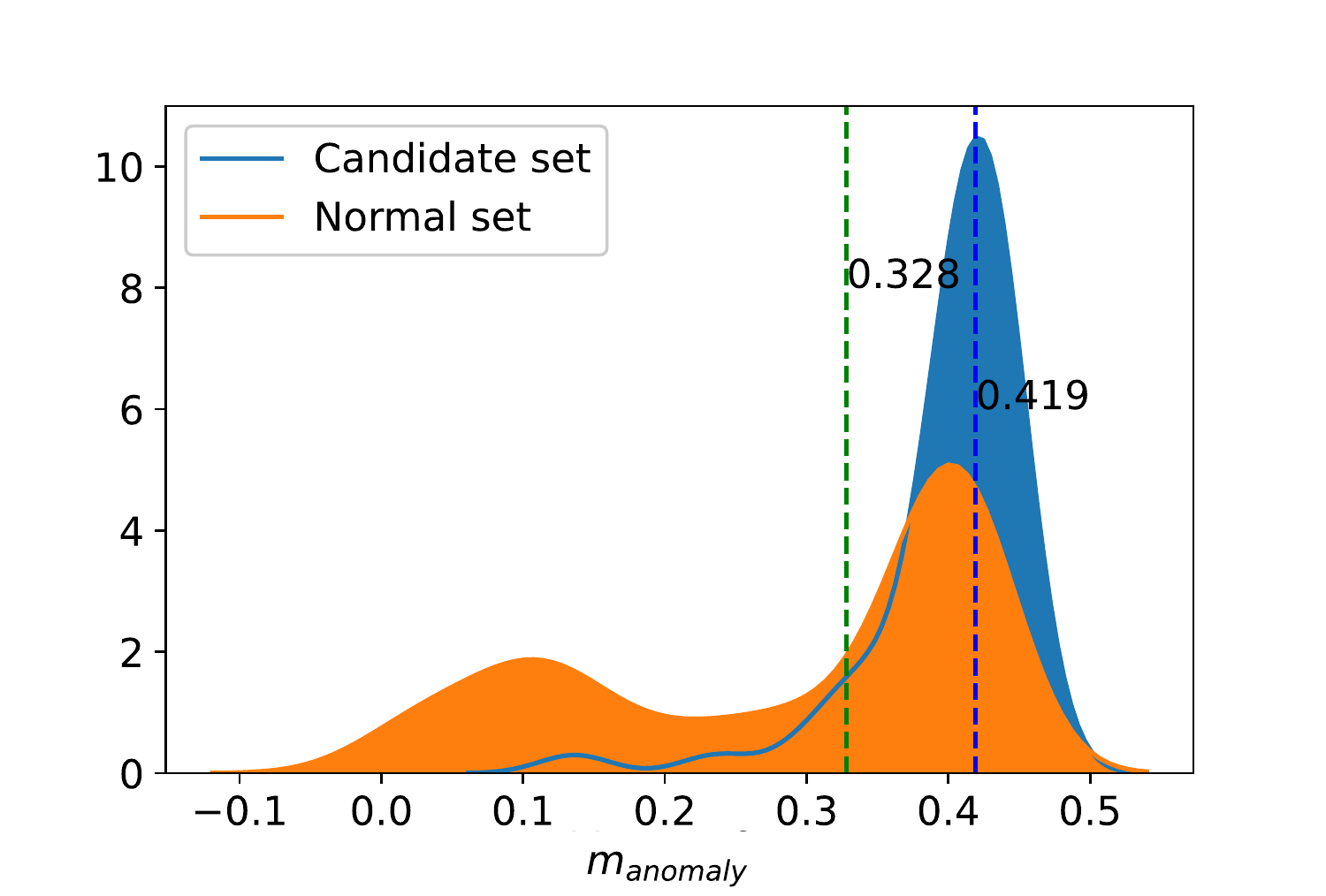}
      \label{fig:CNN_LSTM_Physiological}
      }
    \subfigure[CNN+LSTM -- CAN-Bus Signal]{
     \includegraphics[width=5.5cm]{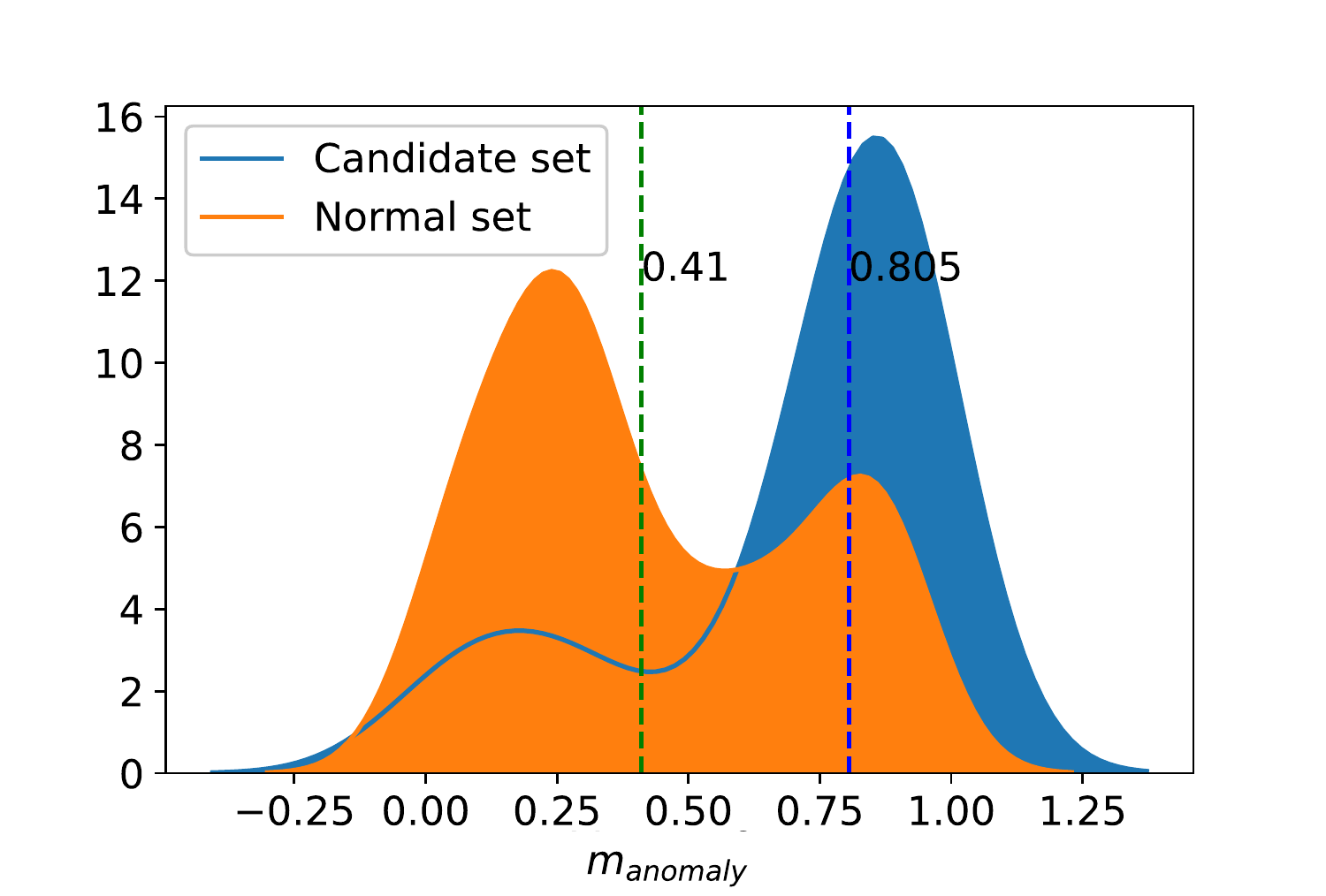}
      \label{fig:CNN_LSTM_CANBUS}
      }
    \subfigure[CNN+LSTM -- All combined]{
     \includegraphics[width=5.5cm]{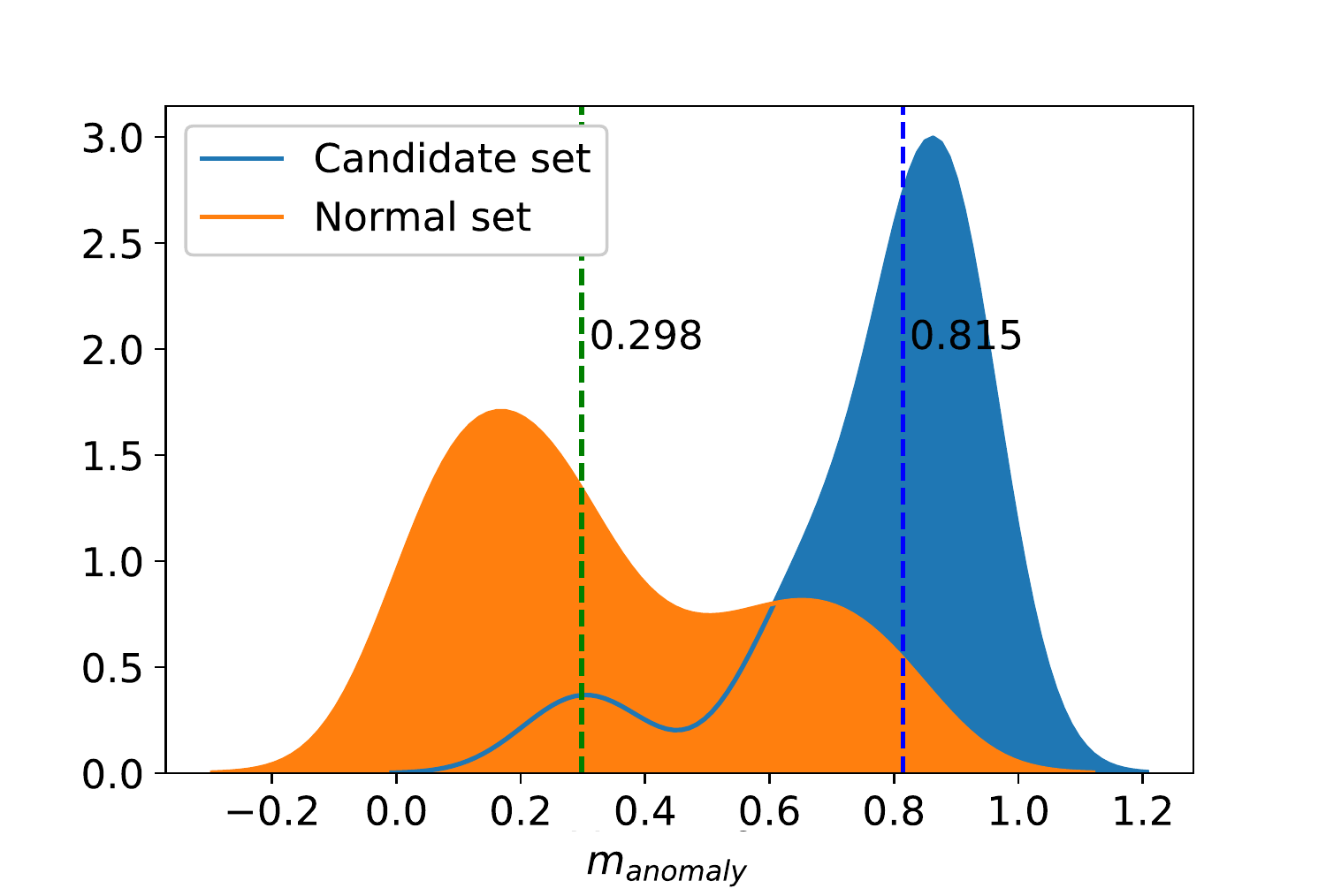}
      \label{fig:CNN_LSTM_ALL}
      }    
  \caption{Histogram of the predictions of the CNN-based, LSTM-based and CNN+LSTM-based models for segments in the test set from the normal and candidate sets  obtained from the DAD corpus. The vertical dashed lines indicate the medians of the anomaly scores for the sets. The results are presented when the models are trained with (1) physiological data, (2) CAN-Bus data, (3), and physiological and CAN-Bus data. }
  \label{fig:CNN_hist}
  \vspace{-0.2cm}
\end{figure*}

Figure \ref{fig:FC_hist} shows the histograms of the anomaly scores for normal and candidate  sets for the FC-based GAN model. The figure shows that the segments from the candidate set have higher anomaly scores than the segments from the normal set. The figure shows clear modes in the histograms showing good separation. Similar results are observed when using  the CNN-based GAN model (Fig. \ref{fig:CNN_ALL}), LSTM-based GAN model (Fig. \ref{fig:LSTM_ALL}), and CNN+LSTM-based GAN model (Fig. \ref{fig:CNN_LSTM_ALL}). In particular, the results for the CNN+LSTM-based GAN model show clear separations, showing the strengths in combining the CNN-based and LSTM-based GAN models, leading to better discriminative performance. We will directly compare all these methods in Section \ref{subsec:DET_curve}.

As described in Section \ref{sec:proposed_methods}, our models are trained with physiological and CAN-Bus signals. We evaluate the contributions of each of these modalities by retraining the models with either physiological or CAN-Bus features. Figure \ref{fig:CNN_hist} shows the histograms for these models. When we only use either physiological or CAN-Bus signals, the differences in the distribution of $m_{anomaly}$ between the normal and candidate sets is clearly reduced, indicating that both modalities provide complementary information. Figures \ref{fig:CNN_CANBUS} and \ref{fig:LSTM_Physiological} are two examples where the overlaps between these distributions are quite clear. Adding both modalities leads to more separation between the distributions, especially for the LSTM-based (Fig. \ref{fig:LSTM_ALL}) and the CNN+LSTM-based (Fig. \ref{fig:CNN_LSTM_ALL}) GAN models. We measure the medians of the distributions to quantify the differences, which are included as vertical dashed lines in the distributions in Figure \ref{fig:CNN_hist}. When the CNN+LSTM-based GAN model is trained with only the physiological signals, the difference between the distributions' medians is $\Delta_{\mathit{Phy}}=0.419-0.328=0.091$ (Fig. \ref{fig:CNN_LSTM_Physiological}). Similarly, when using only CAN-Bus data, the difference between the medians is $\Delta_{\mathit{CAN}}=0.805-0.41=0.395$ (Fig. \ref{fig:CNN_LSTM_CANBUS}). In contrast, the difference between the medians increases to $\Delta_{\mathit{Both}}=0.815-0.298=0.517$ when using both modalities (Fig. \ref{fig:CNN_LSTM_ALL}).

\begin{figure*}[t]
\centering
    \subfigure[Example 1]{
    \includegraphics[width=0.62\columnwidth]{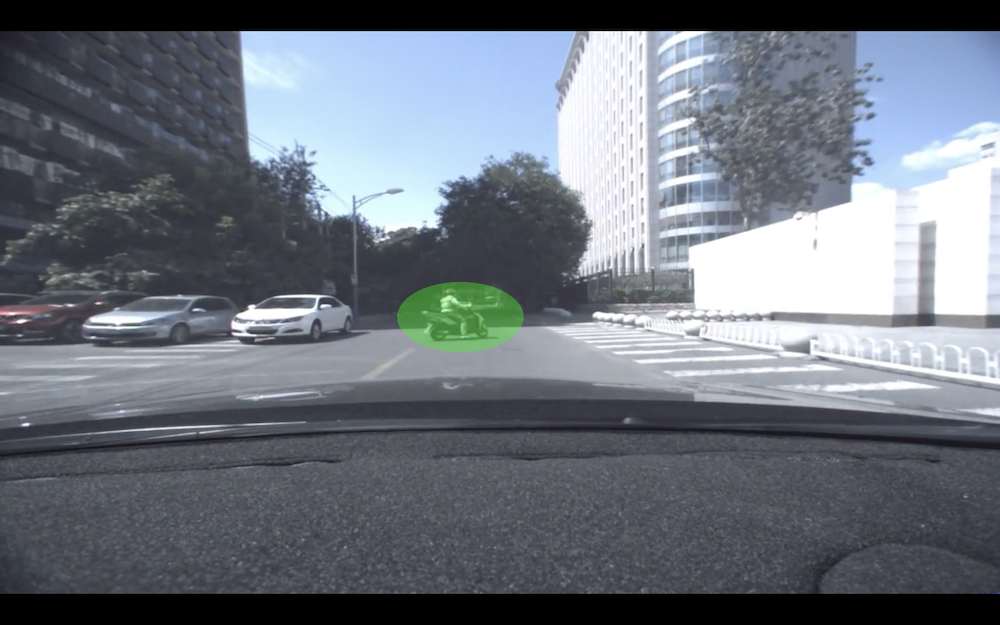}
    \label{fig:examples_a}
    }
    \subfigure[Example 2]{
    \includegraphics[width=0.62\columnwidth]{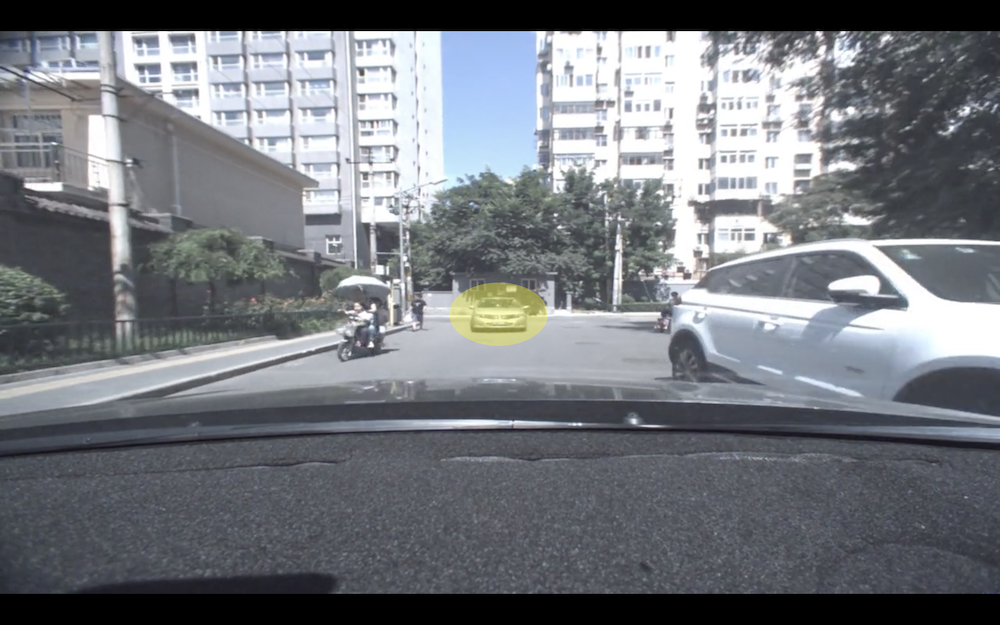}
    \label{fig:examples_b}
    }
    \subfigure[Example 3]{
    \includegraphics[width=0.62\columnwidth]{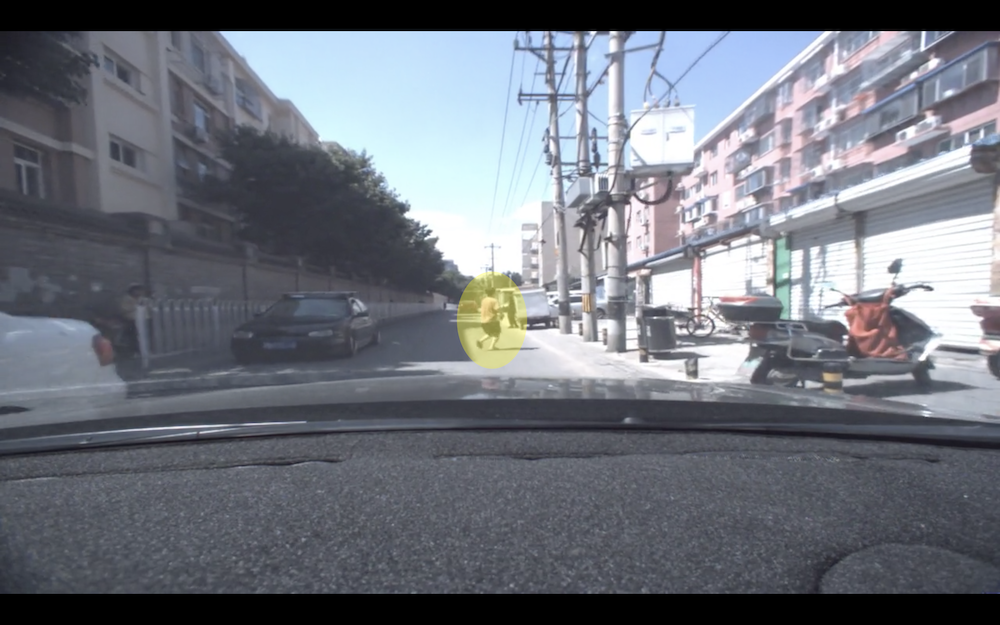}
    \label{fig:examples_c}
    }
    
  \caption{Example of frames from segments with high anomaly scores by the unsupervised CNN+LSTM-based model. (a) A motorcycle suddenly cuts in front of the ego-vehicle, (b) the ego-vehicle is trying to avoid a vehicle parked on the roadside, while a bicyclist is coming in the opposite direction, and (c) a pedestrian suddenly crosses the road, and the driver has to press the brakes to avoid hitting him.}
  \label{fig:examples}
\end{figure*}

Figure \ref{fig:examples} shows some of the abnormal driving scenarios which are discriminated with higher anomaly scores. Most of the anomalies are caused by sudden appearance of pedestrians or improper maneuvers from other vehicles.


\subsection{Direct Comparison of Proposed Models Using DET Curve}
\label{subsec:DET_curve}

We directly compare the models by formulating the evaluation as a binary classification problem (i.e., normal versus candidate sets), where we estimate the \emph{detection error tradeoff} (DET) curves by changing the threshold on the anomaly score. Samples with a score higher than the threshold are classified as abnormal (i.e., part of the candidate set), and those with a score below the threshold are classified as normal. The threshold is not fixed. Instead, we increase its value creating different \emph{operation points}. The DET curves show the \emph{false negative rate} (FNR) in the y-axis and the \emph{false positive rate} (FPR) in the x-axis as the threshold that determines the two classes is moved. The performance of a binary classifier is better when the DET curve lies closer to the axes. The diagonal line in the DET curves indicates \emph{equal error rate} (EER), where the FNR and FPR have the same value. In addition to the DET curves, we also report the EER and the \emph{area under the curve} (AUC) to quantify the results. The AUC is estimated over the DET curves so lower values indicate better performance.

\begin{figure}[t]
\centering
    \includegraphics[width=0.90\columnwidth]{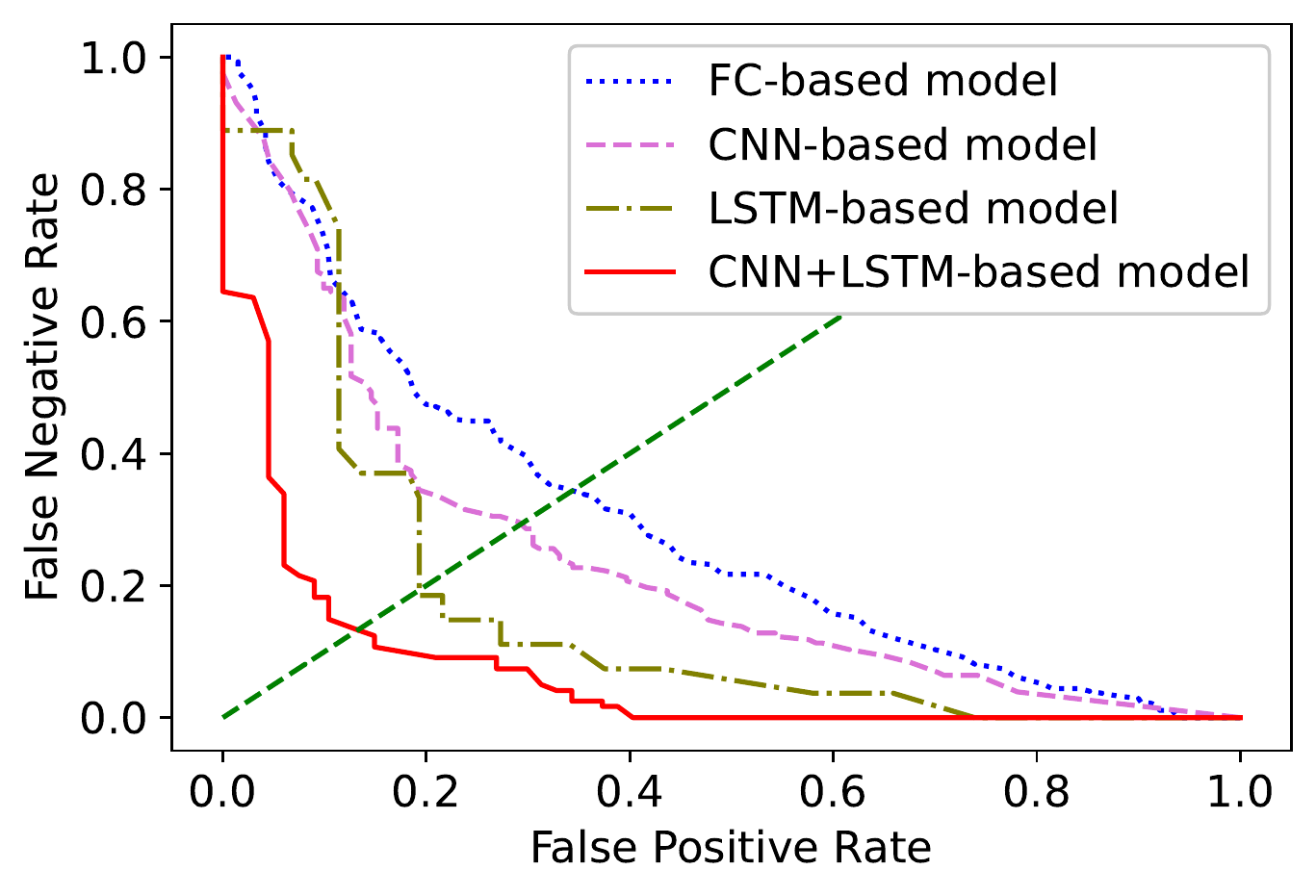}
    \caption{The DET curves for the models by formulating the problem as a binary classification task (candidate versus normal sets). The DET curve shows the false positive rate as a function of the false negative rate. The diagonal dashed line indicates the \emph{equal error rate} (EER) when both error rates are the same.  Generally, the LSTM-based model has better discriminative performance than the CNN-based model and the FC-based model. The best performance is achieved with the CNN+LSTM-based GAN model.}
    \label{fig:ALL_DET}
\end{figure}

Figures \ref{fig:ALL_DET} compares the DET curves of the FC-based, CNN-based, LSTM-based, and CNN+LSTM based GAN models. Table \ref{tab:EERAUC-models} shows the corresponding EER and AUC values. We observe gains in performance over the FC based model by extracting the features directly from the data using CNNs (CNN-based approach), and by modeling temporal information (LSTM-based approach). The performance of the LSTM-based model outperforms the performance of the CNN-based model. This observation is clearly observed in Table \ref{tab:EERAUC-models}. The proposed CNN+LSTM based GAN model achieves the best performance, leveraging the strengths of using CNNs and LSTMs. The CNN+LSTM based GAN model achieves the lowest AUC and EER rates.

\begin{figure}[t]
\centering
\subfigure[CNN-based GAN model]{
 \includegraphics[width=0.8\columnwidth]{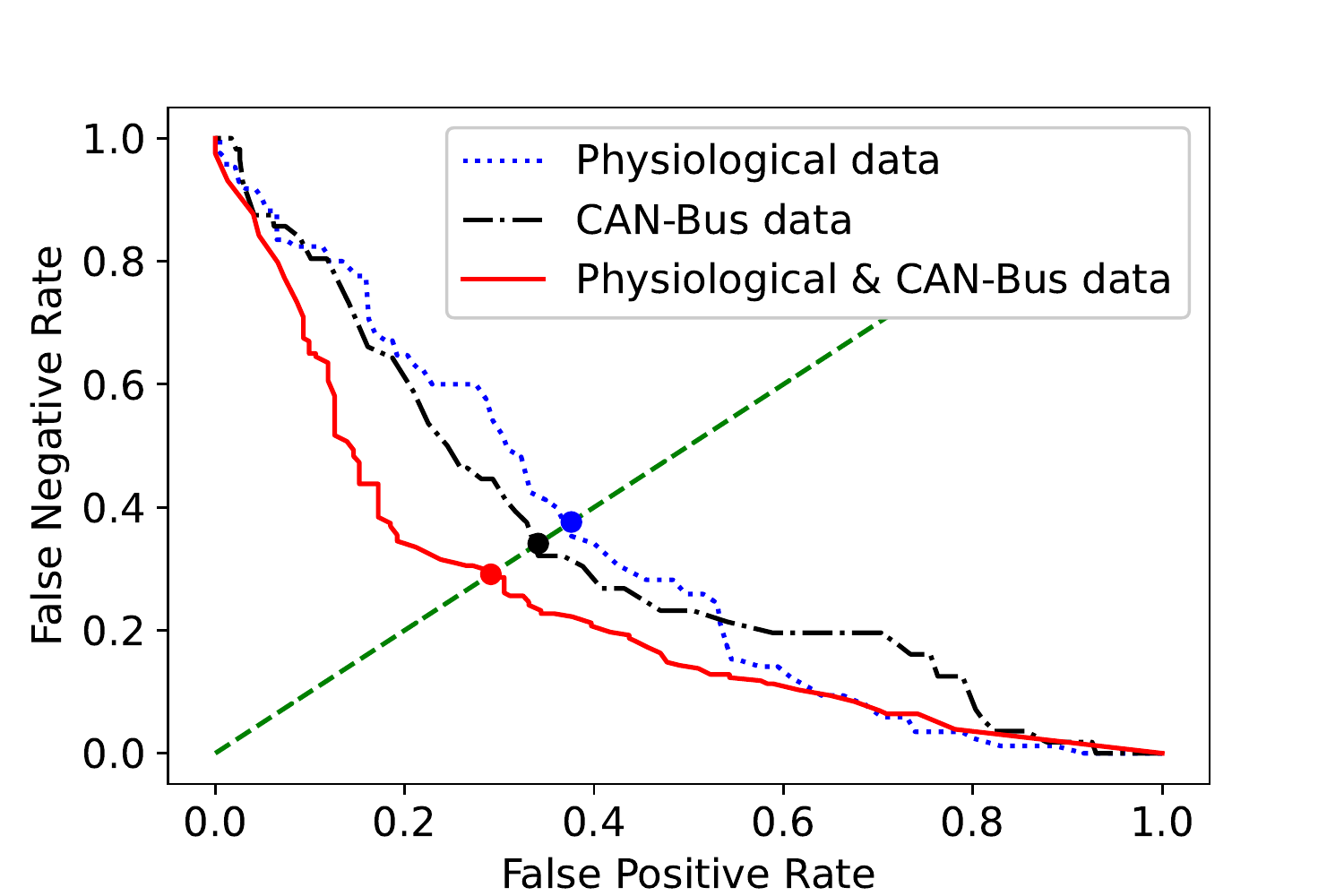}
  \label{fig:CNN_DET}
  }
\subfigure[LSTM-based GAN model]{
 \includegraphics[width=0.8\columnwidth]{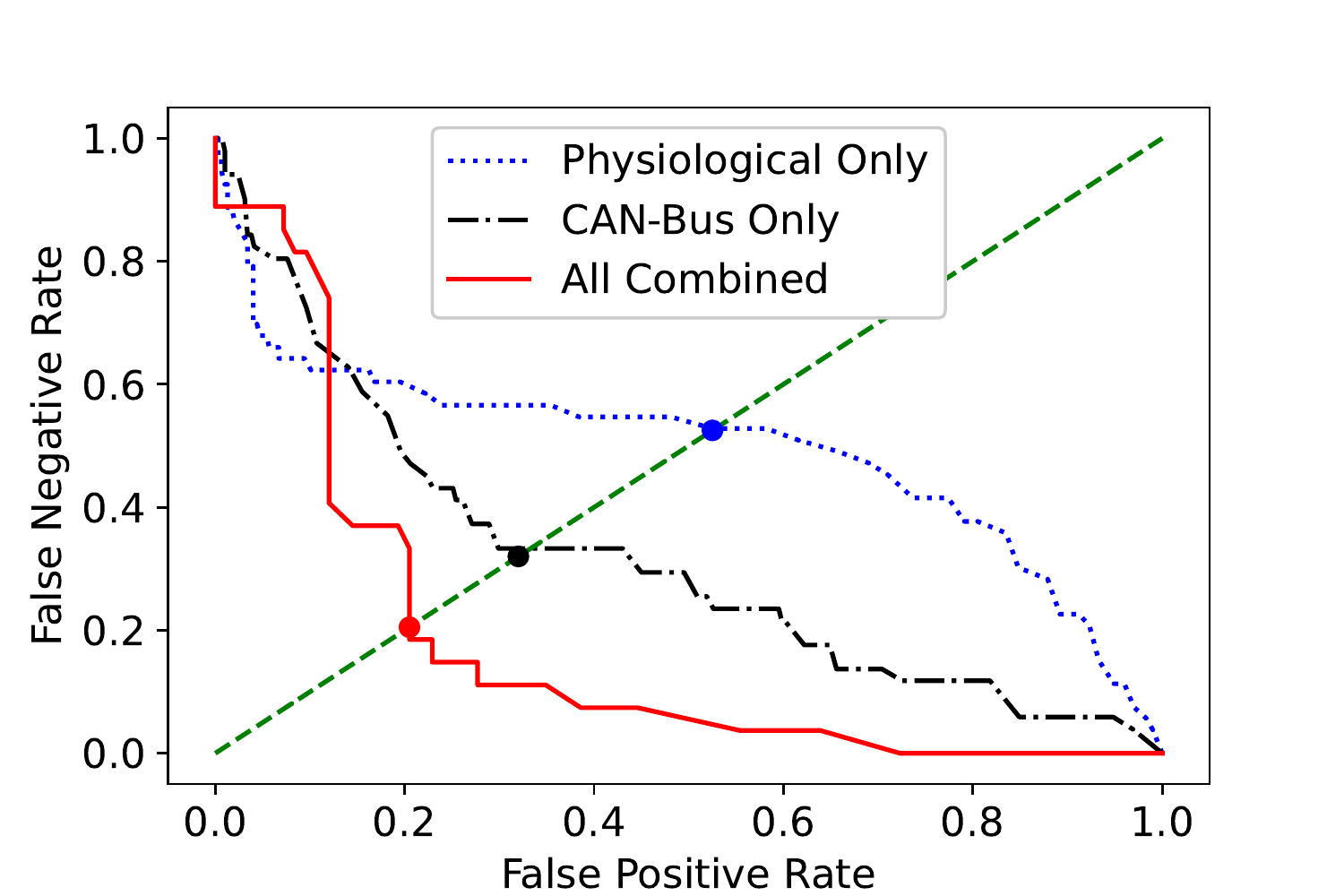}
  \label{fig:LSTM_DET}
  }
\subfigure[CNN+LSTM-based GAN model]{
 \includegraphics[width=0.8\columnwidth]{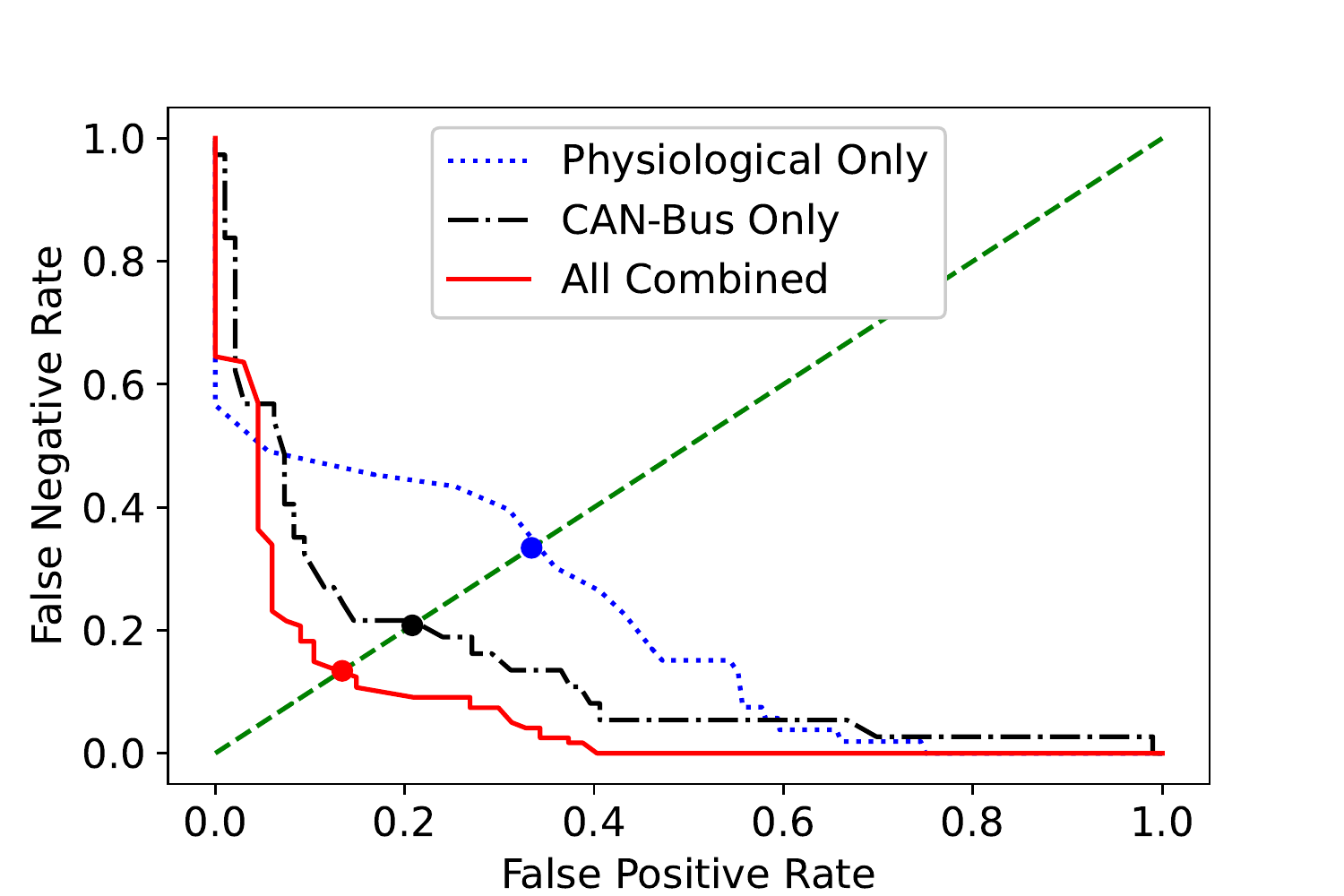}
  \label{fig:CNN_LSTM}
  }
  \caption{The DET curves when the models are trained with (1) physiological data, (2) CAN-Bus data, (3), and physiological and CAN-Bus data. The figure shows that both modalities are important for driving anomaly detection.}
  \label{fig:DET_curves}
\end{figure}

\begin{table}[t]
\centering
\caption{AUC and EER rates for the baseline models and the proposed CNN+LSTM-based model when they are trained with physiological and CAN-Bus signals.}
\fontsize{7.5}{11}\selectfont
  \label{tab:EERAUC-models}
  \begin{tabular}{l@{\hspace{0.1cm}}|@{\hspace{0.1cm}}c@{\hspace{0.2cm}}c@{\hspace{0.1cm}}|c@{\hspace{0.2cm}}c@{\hspace{0.1cm}}|c@{\hspace{0.2cm}}c}
    \hline
    Approach&\multicolumn{2}{c|}{Physiol.} & \multicolumn{2}{c|}{CAN-Bus}& \multicolumn{2}{c}{Both}\\
    &AUC&EER&AUC&EER&AUC&EER\\
    \hline
    \hline
    CNN-based model      & 0.329& 37.6\%& 0.330& 34.1\%& 0.235& 29.1\%\\
    LSTM-based model     & 0.492& 52.5\%& 0.314& 32.0\%& 0.167& 20.5\%\\
    CNN+LSTM-based model & \textbf{0.237}& \textbf{33.4\%}& \textbf{0.176}& \textbf{20.8\%}&\textbf{0.106}& \textbf{13.4\%}\\
  \bottomrule
\end{tabular}
\vspace{-0.2cm}
\end{table}

The DET figures can also be used to compare the results when the models are only trained with  either physiological, or CAN-Bus features. Figure \ref{fig:DET_curves} shows the DET curves for the CNN-based, LSTM-based, and CNN+LSTM based GAN models trained with partial modalities. When we use only physiological data to train the models, we consistently observe lower performance than models only trained with CAN-Bus features. The differences are clearly seen in Table \ref{tab:EERAUC-models} for the AUC and EER rates. However, when we combine physiological and CAN-Bus data, the discriminative performance of the models is improved. These results reveal the significant role of the drivers' physiological data in the performance of our models. This result is also confirmed by observations on the videos with high anomaly scores detected by all the GAN models, when trained with both feature sets. Figure \ref{fig:evaluation_gui} shows a case to illustrate this point, where the driver is slowing down as the vehicle approaches a T-road. All of the sudden, a motorcycle rider rushes into the lane in front of the vehicle. While the driver does not react with any driving maneuver, the driver's breath rate immediately drops, followed by increases in heart rate and skin conductivity. These changes result in a high anomaly score for this driving segment.

\subsection{Comparison with Other Baselines}
\label{subsec:baseline}

We evaluate our approach with four representative baselines inspired by approaches used by previous driving anomaly detection methods. The first baseline corresponds to the \emph{fixed-threshold} method, which is inspired by the work of Li \etal \cite{Li_2016_6}. They detected abnormal driving behaviors by setting thresholds on the vehicle's speed ($VS$), yaw angle ($YA$), acceleration ($AC$) and steering speed ($SS$) data. They defined abnormal speeding and dangerous steering behavior events as:

\begin{align*}
&\mbox{Abnormal speeding} = |AC| > 0.8m/s^{2} \mbox{ \& } SS < 0.1 rad/s\\
&\mbox{Steering =}VS > 30 km/h \mbox{ \&} SS > 0.4 rad/s \mbox{ \&} YA >0.7 rad
\label{eq:abnormal}
\end{align*}

An abnormal driving behavior is defined when either of these two conditions are satisfied. We need to vary the thresholds to compare the performance in DET curves. Our implementation starts with the same threshold as Li \etal \cite{Li_2016_6} for each variable. We consistently move the thresholds across variables by adding or subtracting a fix percentage of their respective range. (i.e., $\hat{t}_{i} = t_{i} + \alpha r_{i}$, where $t_{i}$ is the original threshold for variable $i$, $r_{i}$ is the range of variable $i$, and $\alpha$ is an adjustable parameter to find different tradeoffs between FNR and FPR). The second baseline is the \emph{PCA-threshold} method, which is inspired on the framework proposed by Sadjadi and Hansen \cite{Sadjadi_2013} for \emph{speech activity detection} (SAD). The idea of this unsupervised SAD approach is to combine multiple indicators into a single metric over which we can apply a threshold. This metric corresponds to the first principal component obtained using \emph{principal component analysis} (PCA).  PCA determines the eigenvectors of the covariance matrix of the multidimensional data, which provides the principal directions where the data is spread. The high dimensional feature vectors are linearly mapped into a low dimensional feature space represented by the eigenvector with the highest eigenvalues. We follow the approach presented by Sadjadi and Hansen \cite{Sadjadi_2013} that maps the entire multidimensional data into a single principal dimension. For each 12-second window, we extract the 51-dimensional feature vector from the CAN-Bus and physiological signals discussed in Section \ref{ssec:CGAN-fc}. The vector is mapped into a 1 dimensional metric using this PCA-based approach.  We estimate the DET curve by moving the threshold on the resulting 1-dimensional signal, where the segments with value above the threshold are considered abnormal. We estimate the FPR and FNR rates for different operating points. The third baseline corresponds to the \emph{GMM-threshold} method, which aims to represent the distribution of the data, defining outliers as anomalous events. The \emph{Gaussian mixture model} (GMM) is a common algorithm to fit the distribution of the multidimensional data. A segment can be considered as an outlier if the estimated distribution does not represent well a segment in the test set. Similar to the second baseline, we extracted the 51-dimensional feature vector for each 12-second window, estimating the parameters of a \emph{Gaussian mixture model} (GMM). The number of clusters is set to eight by minimizing the value of the \emph{Akaike information criterion} (AIC) and the \emph{Bayesian information criterion} (BIC). We use the same partitions introduced in Section \ref{sec:data_base} to train and test the GMM.  We calculate the posterior probability of the feature vectors $x$ from the test set according to $p(x) = \sum_{i=1}^{8}\omega _{i}\mathcal{N}(x|\mu _{i}, \sigma _{i}))$, where $\omega _{i}$ (weights), $\mu _{i}$ (mean vector), and $\Sigma _{i}$ (covariance matrix) are the parameters of the GMM. Segments with posterior probability lower than a threshold are considered as outliers (i.e., anomalous events).   The forth baseline is the BeatGAN framework, introduced in Section \ref{ssec:relatedGAN}. We build the BeatGAN model following the description in Zhou \etal \cite{Zhou_2019_2}, setting up the generator in an encoder-decoder structure, using a \emph{multilayer perceptron} (MLP) structure. The numbers of nodes per layer for the generator are 1620-256-128-32-10-10-32-128-256-1620. The numbers of nodes per layer for the discriminator are 1620-256-128-32-1. We calculate the reconstruction error between the real and reconstructed signals as the anomaly score.

\begin{table}[t]
\centering
\caption{AUC and EER rates for GAN-based models when they are either trained with physiological signals, CAN-Bus signals or physiological and CAN-Bus signals.}
  \label{tab:EERAUC-baseline}
  \begin{tabular*}{0.9\columnwidth}{@{\extracolsep{\fill}}l|cc}
    \hline
    Approach&\multicolumn{2}{c}{Physiological \& CAN-Bus}\\
    &AUC & EER\\
    \hline
    \hline
    Fixed threshold	        &0.433	&45.1\%\\
    GMM threshold   	    &0.341	&38.6\%\\
    PCA-threshold	        &0.312	&39.4\%\\
    BeatGAN threshold   	&0.252	&31.1\%\\
    \hline
    CNN+LSTM-based model	&\textbf{0.106}	&\textbf{13.4\%}\\
  \bottomrule
    \end{tabular*}
\vspace{-0.2cm}
\end{table}

\begin{figure}[t]
\centering
    \includegraphics[width=0.90\columnwidth]{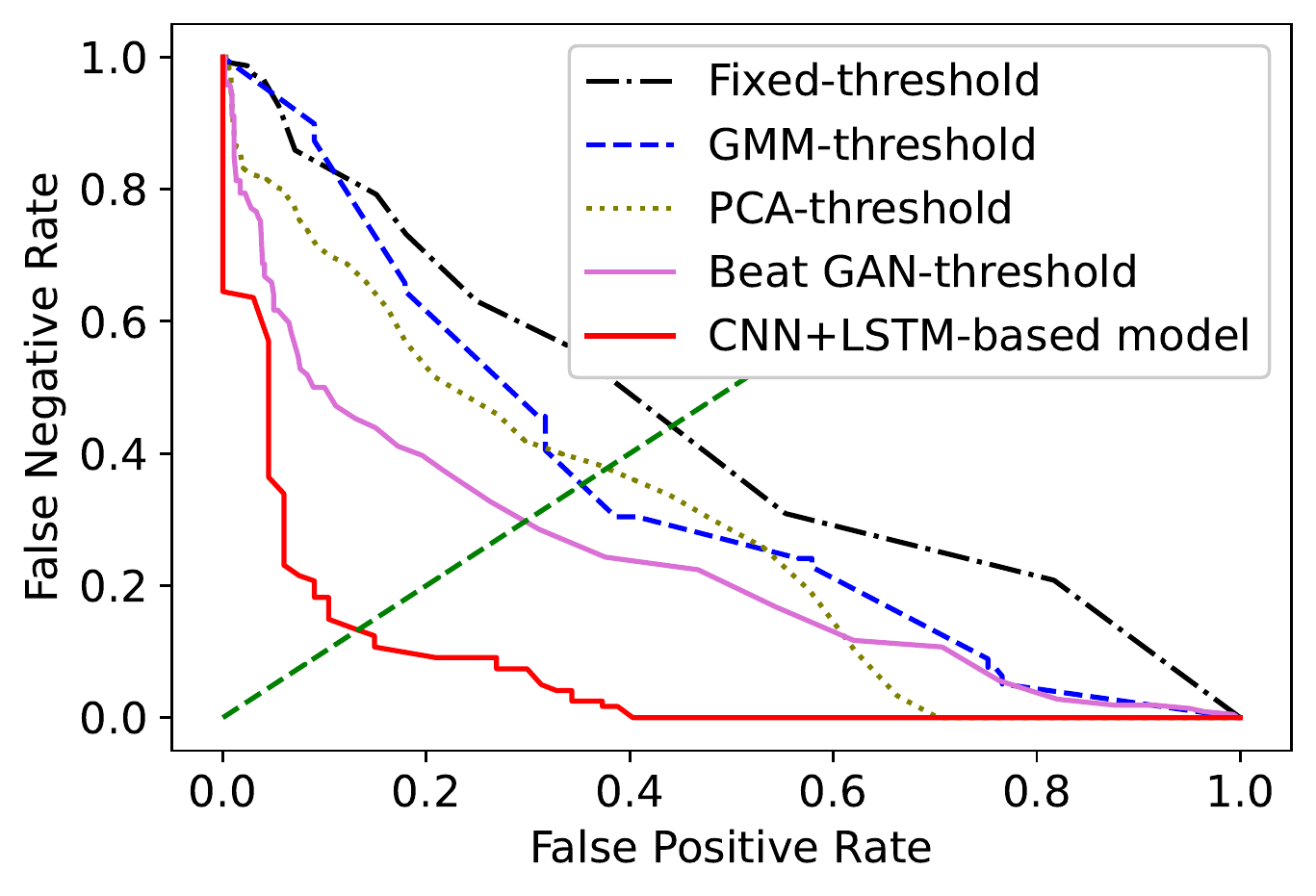}
    \caption{The DET curves comparing the proposed models with the four baseline models. The figure shows that the proposed model outperforms the baseline methods.}
    \label{fig:DET_baseline}
\end{figure}

Figure \ref{fig:DET_baseline} shows the DET curves comparing the baseline methods with our CNN+LSTM based model. Table \ref{tab:EERAUC-baseline} shows the corresponding EER and AUC results. The figure shows that our proposed model leads to clear improvements over the baseline models. The results show that the BeatGAN framework is the best baseline model. However, our proposed approach clearly outperforms this method.

\subsection{Overlap Between Selected Segments and DAD Annotations}
\label{subsec:annotation}

This section discusses the results by evaluating the annotations in the DAD corpus overlapping with the driving segments with the highest anomaly scores in the test set. 
Successful anomaly detection models should identify video segments belonging to the  candidate set. If this is not the case, we expect to observe driving maneuvers which may also correlate with anomaly scenarios. For each model, we identify the top 100 segments with the highest anomaly scores. Then, we count the number of annotations overlapping with these segments, grouping the annotations into normal, candidate or maneuver (Table \ref{tab:sets}). As a baseline of this evaluation, we randomly select 100 six-second videos from the test set. We compare the FC-based, CNN-based, LSTM-based, and CNN+LSTM based GAN models implemented with physiological and CAN-Bus data.

\begin{table}[t]
\caption{Overlap between the top 100 segments selected by the proposed models and the DAD annotations. The table also includes 100 segments randomly selected. We consider the normal, candidate, and maneuver sets (Table \ref{tab:sets}).}
  \label{tab:score_distribution}
  \begin{tabular}{l|ccc}
    \hline
    Model & Normal & Candidate & Maneuver\\
    \hline
    \hline
    Random & 54 & 4 & 42\\
    FC-based model  & 31 & 9 & 60 \\   
    CNN-based model & 27 & 10 & 63\\
    LSTM-based model & 19 & 16 & 65\\
    CNN+LSTM-based model & 11 & 21 & 68\\
  \bottomrule
\end{tabular}
\vspace{-0.2cm}
\end{table}

Table \ref{tab:score_distribution} shows the results. Most of the videos in the random set are from the normal set (i.e., 54\%). Only 4\% of the videos are included in the candidate set. In contrast, the proportion of normal segments identified by the proposed models significantly drops (11\%-31\%). Most of the selected segments are either from the candidate set, or are associated with driving maneuvers. The table shows that the number of segments from the candidate set increases up to 21\%. The LSTM-based model also identifies more videos from the candidate and maneuver sets than the CNN-based model. The discriminative performance of the CNN+LSTM-based model is better than the LSTM-based and CNN-based models, confirming the observation made from the histograms (Fig. \ref{fig:CNN_hist}) and DET curves (Fig. \ref{fig:ALL_DET}). These results indicate that our unsupervised approach is effective in identifying driving segments of interest.

\subsection{Perceptual Evaluation}
\label{subsec:perceptual_evaluation}

\begin{figure}[t]
  \centering
  \includegraphics[width=0.90\columnwidth]{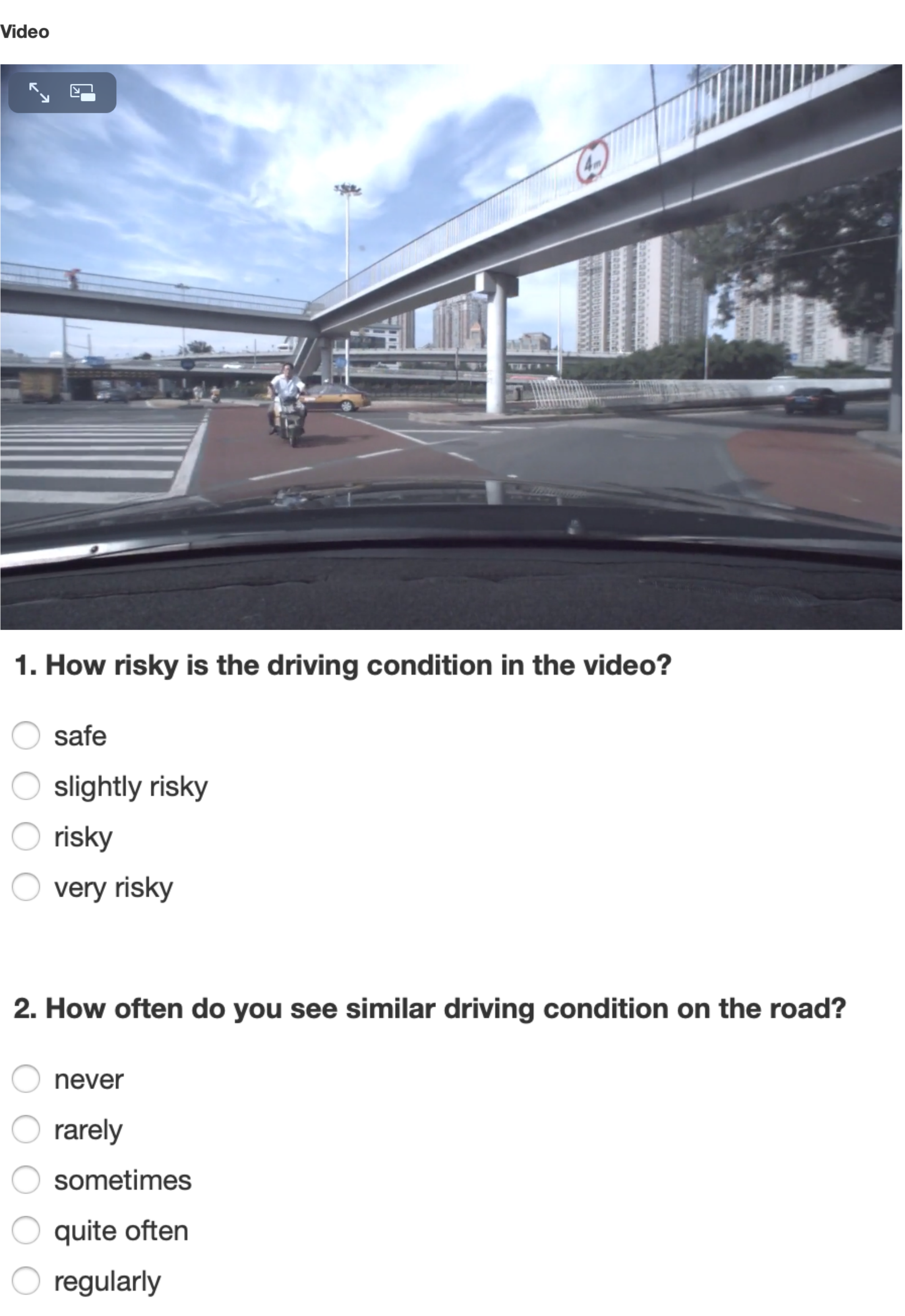}
  \caption{Graphical user interface for the perceptual evaluation of driving anomaly. After watching each video, the raters are asked to assess the level of risk and familiarity in the segment.}
  \label{fig:evaluation_gui}
  \vspace{-0.2cm}
\end{figure}

In  previous sections, we evaluated the models with the annotations of the DAD corpus. However, these annotations do not directly indicate the anomalous level of the segment. This section evaluates the discriminative results using perceptual evaluations, directly assessing the level of anomaly in the selected segments. We select 100 driving segments with the highest anomaly scores for the CNN-based, LSTM-based, and CNN+LSTM based GAN models. We also randomly select 100 videos from the test set. To provide enough context, the duration of each segment is 12 seconds, containing the six-second segment used as condition and the six-second segment of the data predicted by the models. We only have 285 unique segments to evaluate due to the overlap of the driving segments selected by the different models. Figure \ref{fig:evaluation_gui} shows the \emph{graphical user interface} (GUI) for the evaluation. Nine raters participated in the evaluation process, where each of them assessed 95 different videos. The evaluators are university students, where seven are male and two are female. Their average age at the time of the evaluation was 24.7 years old. As a result, each video is evaluated three times by three different raters. After watching each video, the raters are asked to answer two questions: (1) \emph{how risky is the driving condition in the video? (safe; slightly risky; risky; very risky)}, and (2) \emph{how often do you see similar driving condition on the road? (never; almost never; sometimes; quite often; regularly)}. These two questions aim to assess the degree of risk and familiarity of the driving conditions shown in the videos. We estimate the inter-rater reliability among the nine raters who participated in the perceptual evaluation. The annotators were grouped into three groups, so that each of the driving segments is evaluated by three raters. Since the two questions in the evaluations are Likert-like scales, we use the Krippendroff's alpha coefficient to assess the inter-evaluator agreement. Table \ref{tab:interRater} shows the results. We observe a substantial agreement for question 1 (i.e., risky level) and moderate agreement for question 2 (i.e., familiarity level). The agreement levels are consistent across the three groups.

\begin{table}[t]
\centering
\caption{Krippendroff's alpha coefficients among different groups of raters. Each group has three raters who assessed the same videos. The questions correspond to the questionnaire presented in Figure \ref{fig:evaluation_gui}.}
  \label{tab:interRater}
  \begin{tabular}{l|cccc}
    \hline
    & Group 1 & Group 2 & Group 3 & Over All \\
    \hline
    \hline
    Question 1 & 0.745 & 0.711 & 0.706 & 0.736\\
    Question 2 & 0.512 & 0.532 & 0.653 & 0.573\\
  \bottomrule
\end{tabular}
\vspace{-0.2cm}
\end{table}

\begin{figure}[t]
\centering
\subfigure[How risky is the driving condition in the video?]{
  \includegraphics[trim=3.6cm 0.7cm 3.8cm 1.8cm, clip,width=0.98\columnwidth]{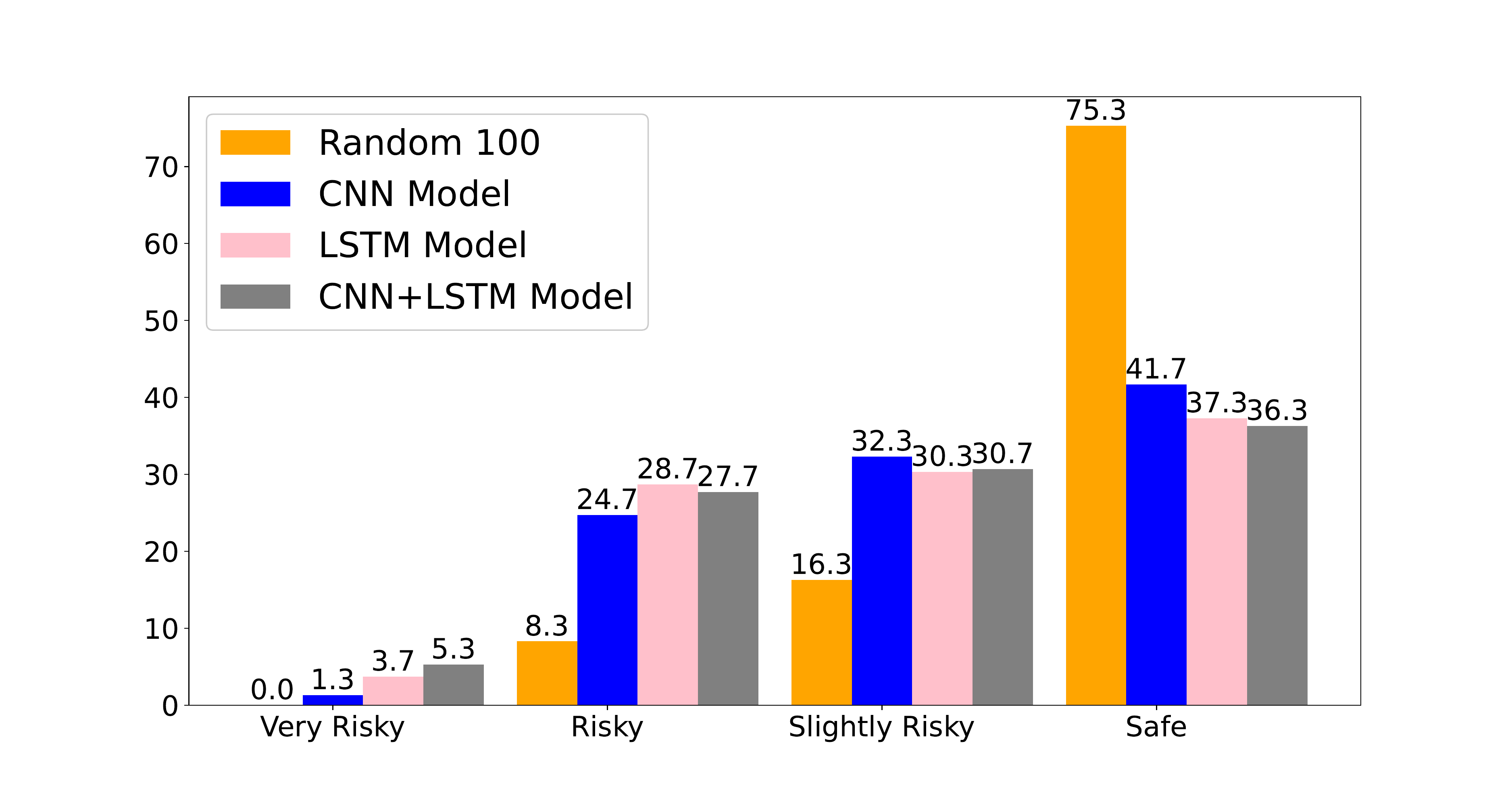}
    \label{fig:perception_evaluation_a}
  }\\ 
\subfigure[How often do you see similar driving condition on the roads?]{
  \includegraphics[trim=3.6cm 0.7cm 3.8cm 1.8cm, clip,width=0.98\columnwidth]{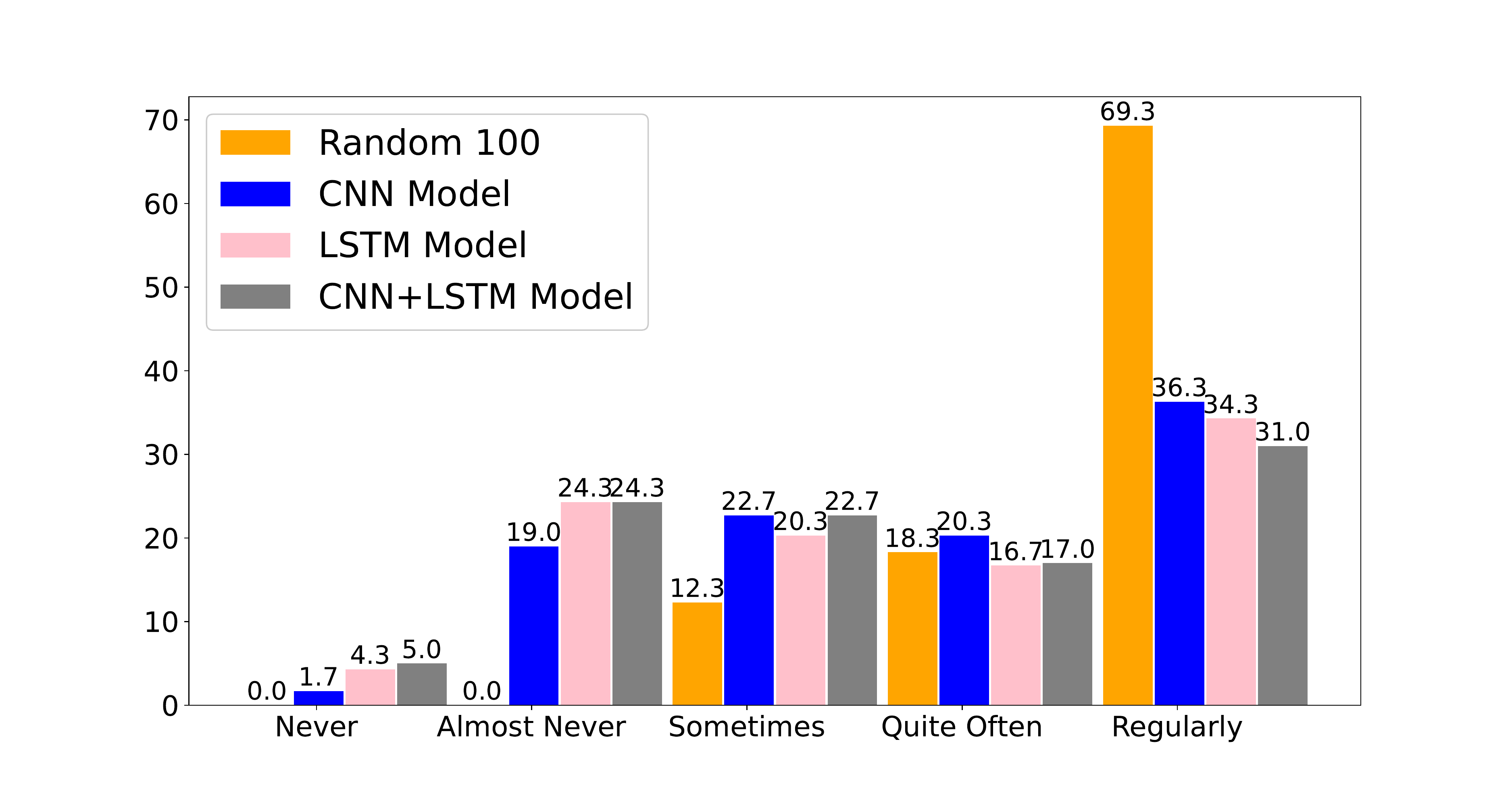}
    \label{fig:perception_evaluation_b}
  }
  \caption{Results of the perceptual evaluation to assess the degree of risk and familiarity of the selected videos. The figures show the results for the top 100 segments with the highest anomaly scores selected by three different models. The figure also shows the results for 100 segments randomly selected.}
  \label{fig:perception_evaluation_result}
  \vspace{-0.2cm}
\end{figure}

Figure \ref{fig:perception_evaluation_result} shows the evaluation results.  Each model independently selects 100 segments with the highest scores. Since the methods are different, the selected videos differ across methods, where some of the videos overlap across methods. The differences in the selected sets are reflected in the differences observed in Figure \ref{fig:perception_evaluation_result}. A video is annotated by three raters. Therefore, for each model, we have 300 annotations assigned by the raters. Then, we determine the proportion of the annotations assigned to each of the options listed in the GUI (Figure \ref{fig:evaluation_gui}), providing the results in Figure \ref{fig:perception_evaluation_result}. For example, the number of annotations assigned to each of the options for the question ``How risky is the driving condition in the video?'' for the CNN+LSTM model are: \emph{very risky} 16 (5.3\%), \emph{risky} 83 (27.7\%), \emph{slightly risky} 92 (30.7\%), and \emph{safe} 109 (36.3\%). Figure \ref{fig:perception_evaluation_result} shows that the videos selected by the CNN+LSTM based GAN model contain more segments annotated with higher risk and lower familiarity. The figure shows that 33\% of these videos are considered as \emph{risky} or \emph{very risky}, and 29.3\% of them are considered to occur \emph{never} or \emph{almost never}. The corresponding percentages for the CNN-based and LSTM-based models are lower. These comparisons illustrate that the CNN+LSTM-based model can identify more hazardous and more abnormal driving conditions than the CNN-based model and the LSTM-based model. These numbers are significant, since 75\% of the randomly selected segments are considered safe, and 69.3\% of them are considered to occur \emph{regularly}, showing that most of the driving conditions in the DAD corpus are regular scenarios without driving anomalies. Our best unsupervised conditional GAN model is able to identify segments which are often perceived with a level of risk (64.7\%), which do not occur so often on the road. The figure also shows the superior performance of the LSTM-based model compared to the CNN-based models, validating the results observed in previous sections with the DAD annotations.



\section{Conclusions}
\label{sec:conclusion}


This study proposed an unsupervised driving anomaly detection system based on a conditional GAN. The proposed approach makes predictions of the driver's physiological data and the vehicle CAN-Bus data, conditioning the model on previous observed signals. The predictions are contrasted with actual data, creating an anomaly score that increases its value when unexpected data is observed. The approach obtains discriminative features from the physiological and CAN-Bus signals directly from the data using CNNs. The model also leverages temporal information by using LSTM networks. This study shows that the driving events with more hazardous driving conditions usually receive higher anomaly scores by the proposed model. This result is validated with objective evaluations, relying on the annotations of the DAD corpus, and perceptual evaluations conducted on the videos selected by our models with the highest anomaly scores. Our proposed approach is able to effectively detect anomaly driving conditions that deviate from the predictions of the upcoming driving behaviors, creating an appealing unsupervised solution that does not depend on either predefined thresholds or supervised rules.

One limitation of this study is that our detection algorithm depends on actions or reactions from the driver. Anomalies can be detected only when the driver notices events and reacts to them. Therefore, if the driver is unaware of a driving anomaly, a model trained with physiological and CAN-Bus signals will not provide discriminative information to detect it. A potential solution is adding other features that can objectively capture the driver's  environment regardless of her/his awareness (e.g., pedestrian detection, car detection).  We plan to augment our proposed model with further information, such as the results from vision-based object detection systems. Another limitation of our approach is the use of wearable devices to capture physiological signals. Future research directions include developing remote approaches to measure the driver's physiological signals \cite{Yu_2019, Yu_2019_2}. While today this is a challenging problem, there are technological advances to create non-contact measurement systems to monitor physiological data that suggest that this could be reasonable in the future. Sensors can be installed on the driver's seat to record the driver's BR and HR signals. Physiological signals can be alternatively obtained from wearable sensors that the drivers may be already using (e.g., smartwatch). Development in this area will make our approach more suitable for deployment in real-driving conditions.


%





\ifCLASSOPTIONcaptionsoff
  \newpage
\fi



%

\bibliographystyle{IEEEtran}
\bibliography{reference}

\end{document}